%% file: arxiv.tex
\documentclass{article}

\usepackage[utf8]{inputenc} 
\usepackage[T1]{fontenc}    
\usepackage{hyperref}       
\usepackage{url}            
\usepackage{booktabs}       
\usepackage{amsfonts}       
\usepackage{nicefrac}       
\usepackage{microtype}      
\usepackage{xcolor}         

\usepackage{mathrsfs}
\usepackage{amsmath}
\usepackage{amsthm}
\usepackage{amssymb}
\usepackage{mathtools}
\usepackage{bm}

\usepackage{babel}
\usepackage{color}
\PassOptionsToPackage{round,semicolon}{natbib}
\usepackage{hyperref}
\hypersetup{colorlinks=true,citecolor=blue,linkcolor=blue,urlcolor=blue}

\usepackage{graphicx}
\usepackage{caption}
\usepackage{subcaption}

\usepackage{booktabs}
\usepackage{longtable}
\usepackage{xurl}
\usepackage{array}
\usepackage{multirow}
\newcolumntype{C}[1]{>{\centering\arraybackslash$}m{#1}<{$}}

\newtheorem{theorem}{Theorem}
\newtheorem{corollary}{Corollary}

\usepackage[main, final]{neurips_2026}
\title{Affine-Image Propagation for Tight and Scalable Neural Network Verification}

\author{%
\begin{NoHyper}
  Hong-Ming Chiu\thanks{Equal contribution} \end{NoHyper} \\
  University of Illinois at Urbana--Champaign\\
\texttt{hmchiu2@illinois.edu}\\
\And 
\begin{NoHyper}
Haoyu Li\footnotemark[1]
\end{NoHyper} \\
University of Illinois at Urbana--Champaign\\
\texttt{haoyuli5@illinois.edu}
\And Huan Zhang\\
University of Illinois at Urbana--Champaign\\
\texttt{huan@huan-zhang.com}
 \And Richard Y. Zhang\\
University of Illinois at Urbana--Champaign\\
\texttt{ryz@illinois.edu}
}

\usepackage{xcolor}

\begin{document}
\maketitle
\global\long\def\zlb{\underline{z}}%
\global\long\def\zub{\overline{z}}%
\global\long\def\xlb{\underline{x}}%
\global\long\def\xub{\overline{x}}%
\global\long\def\pub{\overline{p}}%
\global\long\def\qub{\overline{q}}%
\global\long\def\R{\mathbb{R}}%
\global\long\def\S{\mathbb{S}}%
\global\long\def\XX{\mathcal{X}}%
\global\long\def\BB{\mathcal{B}}%
\global\long\def\PP{\mathcal{P}}%
\global\long\def\E{\mathcal{E}}%
\global\long\def\ZZ{\mathcal{Z}}%
\global\long\def\inner#1#2{\left\langle #1,#2\right\rangle }%

\global\long\def\1{\mathbf{1}}%
\global\long\def\g{\mathbf{g}}%
\global\long\def\blambda{\bm{\lambda}}%
\global\long\def\btau{\bm{\tau}}%

\global\long\def\R{\mathbb{R}}%
\global\long\def\X{\mathcal{X}}%
\global\long\def\T{\mathcal{T}}%
\global\long\def\relu{\operatorname{ReLU}}%
\global\long\def\conv{\operatorname{conv}}%
\global\long\def\diag{\operatorname{diag}}%
\global\long\def\tr{\operatorname{tr}}%
\global\long\def\dd{\operatorname{dd}}%
\global\long\def\BP{\mathrm{BP}}%
\global\long\def\LP{\mathrm{LP}}%
\global\long\def\SDP{\mathrm{SDP}}%
\global\long\def\one{\mathbf{1}}%
\global\long\def\cond{\operatorname{cond}}%
\global\long\def\ip#1#2{\left\langle #1,#2\right\rangle }%

\maketitle
\begin{abstract}
While bound propagation methods are highly effective for certifying neural networks against $\ell_{\infty}$ adversaries, tight and scalable verification for $\ell_{2}$ adversaries remains a significant open challenge. Existing methods incur geometric loss by enclosing preactivation sets in elementwise intervals or axis-aligned ellipsoids, and functional loss by restricting the linear bounds constructed over those domains. Consequently, tight $\ell_{2}$ verification is currently restricted to Lipschitz-regularized networks where these approximations are accurate. In this paper, we introduce affine-image SDP-CROWN, a novel bound propagation method that better approximates preactivation domains using affine-image constraints and jointly optimizes the linear bound to adapt to the resulting domain. To bypass the prohibitive computational cost of dense semidefinite constraints required by the  affine-image representation, we replace them with scaled diagonal-dominance conditions, allowing a GPU-accelerated implementation. Our method preserves the bound propagation recursion and can be integrated with branch-and-bound. On five MNIST and CIFAR-10 non-Lipschitz-regularized networks, our method improves verified accuracy over state-of-the-art methods by 17\% to 57.5\%. With branch-and-bound, these improvements increase to between 27\% and 63.5\%. The code for reproducing our experimental results is available at~\url{https://github.com/Verified-Intelligence/Affine-Image-SDP-CROWN}.
\end{abstract}

\section{Introduction}

Bound propagation \citep{zhang2018crown,singh2018fast,singh2019abstract,wang2018efficient,dvijotham2018dual,hashemi2021osip,xu2020automatic,xu2021fast} has emerged as a highly effective technique for
the formal verification of neural networks against $\ell_{\infty}$
adversaries, yet it surprisingly struggles when faced with $\ell_{2}$
adversaries. Addressing this disparity is increasingly critical. Modern robustness questions are moving beyond pixel-wise perturbations in image space toward semantic perturbations in latent representations \citep{wong2021learning,barrett2022certifiably}, where Euclidean geometry is the natural model. Scalable $\ell_{2}$ verification is therefore a necessary step toward certifying modern representation learners.

Existing bound propagation methods can incur two sources of looseness: geometric loss from approximating preactivation sets by larger, simpler domains, and functional loss from restricting the linear bounds constructed over those domains. For $\ell_2$ perturbations, the reachable preactivation set is obtained by mapping the input Euclidean ball through the network. Methods based on elementwise intervals, such as $\alpha$-CROWN \citep{xu2021fast}, construct their linear activation bounds over coordinatewise preactivation intervals. These bounds must therefore remain valid for combinations of preactivations outside the reachable set. For example, the elementwise box of an $n$-dimensional ball has a Euclidean radius $\sqrt{n}$ times that of the ball. Constructing activation bounds over such enlarged domains can introduce geometric loss and weaken the resulting certificates.

A significant breakthrough in reducing this geometric loss came from
SDP-CROWN~\citep{chiu2025sdp}, which integrated semidefinite programming
(SDP) relaxations~\citep{raghunathan2018semidefinite,brown2022unified} into scalable bound propagation. Although SDP relaxations have long provided tight certificates for $\ell_2$ verification~\citep{zhang2020tightness,chiu2023tight}, their computational cost has limited their scalability. SDP-CROWN addressed this bottleneck by deriving layerwise linear bounds from SDP relaxations and incorporating them into the backward pass, preserving the computational structure of bound propagation. The key insight is that, for preactivation domains modeled as Euclidean balls or axis-aligned ellipsoids, the SDP relaxation yields an explicit family of linear bounds. These bounds incorporate joint $\ell_2$ constraints directly into bound propagation, avoiding the $\sqrt{n}$ increase in Euclidean radius that arises when a ball is enclosed in an elementwise box.

However, SDP-CROWN \citep{chiu2025sdp}  relies on the propagated axis-aligned ellipsoids remaining informative throughout the network. Multiplication by a weight matrix generally changes both the orientation of an ellipsoid. To maintain efficiency, SDP-CROWN encloses the resulting image using another axis-aligned ellipsoid. This approach works well on the Lipschitz-regularized benchmarks, where even naive spectral-norm propagation provides informative bounds. For non-Lipschitz-regularized networks, such enclosures can substantially enlarge the set along directions in which the image is narrow, introducing geometric loss. The resulting SDP constraints may then be too loose to yield tight certificates, limiting the effectiveness of SDP-CROWN on these networks.

This leaves a fundamental research question: \emph{How can we achieve tight verification against $\ell_2$ adversaries on non-Lipschitz-regularized networks while retaining the scalability of bound propagation?} To address this question, we develop affine-image SDP-CROWN, which augments elementwise interval and axis-aligned ellipsoid constraints with the affine image of a previous-layer outer approximation. For an ellipsoidal $\ell_2$ input set, this representation captures the first preactivation domain exactly. In subsequent layers, it preserves the affine geometry of the previous-layer outer approximation, with approximation error inherited from earlier layers. We jointly optimize the slopes and offsets of the propagated linear bounds to exploit these richer domains.

The main technical challenge is to construct linear bounds over affine-image domains efficiently. The associated SDP relaxation introduces a dense positive semidefinite constraint, whose direct enforcement can be computationally expensive. We derive tractable sufficient conditions based on scaled diagonal-dominance and optimize the scaling jointly with the bound parameters. For convolutional layers, our GPU implementation exploits sparsity to evaluate the required terms and gradients without forming dense matrices. The resulting certificates incorporate affine-image constraints while preserving the linear bound propagation recursion, making stronger $\ell_2$ verification practical for non-Lipschitz-regularized networks.

\subsection{Contributions}

We introduce affine-image SDP-CROWN for scalable $\ell_2$ verification of non-Lipschitz-regularized ReLU networks. Our method augments interval and ellipsoidal constraints with affine images of previous-layer outer approximations and jointly optimizes the propagated linear bounds, addressing both geometric and functional loss. Scaled diagonal-dominance conditions and a GPU implementation that exploits convolutional sparsity make the resulting certificates tractable. The method preserves the bound propagation recursion, integrates with branch-and-bound, and substantially improves verified accuracy on the evaluated MNIST and CIFAR-10 networks.

\subsection{Related work}
Semidefinite programming (SDP) relaxation is a principled approach for verifying neural networks against $\ell_2$ adversaries, due to its ability to capture the geometry induced by $\ell_2$-norm. Consequently, SDP-based methods \citep{raghunathan2018semidefinite,fazlyab2019efficient,batten2021efficient,ma2020strengthened,fazlyab2020safety,dvijotham2020efficient} provide some of the tightest known verification against $\ell_2$ adversaries. However, SDP relaxations suffer from poor scalability; even with recent advances in solver design \citep{chiu2023tight} and hardware acceleration \citep{dathathri2020enabling}, they remain computationally prohibitive for networks with more than 10 thousand neurons. Moreover, their high computational cost prevents the propagation of intermediate preactivation bounds using the same SDP relaxation. As observed in \citep{raghunathan2018semidefinite,dathathri2020enabling,chiu2025sdp}, tight intermediate bounds are critical for maintaining the effectiveness of SDP relaxations in deep networks. 

Alternatively, bound propagation methods based on linear programming (LP) relaxations \citep{zhang2018crown,singh2018fast,singh2019abstract,wang2018efficient,dvijotham2018dual,hashemi2021osip,xu2020automatic,xu2021fast} enable efficient and highly parallelizable computation of intermediate bounds. While these methods are primarily designed for $\ell_\infty$ adversaries and are generally looser than SDP-based approaches, they scale to networks with millions of neurons. To improve verification tightness, prior works combine bound propagation with branch-and-bound \citep{wang2021beta,de2021scaling,ferrari2022complete,shi2024genbab}, cutting planes \citep{zhang2022general,zhou2024scalable}, and domain clipping techniques \citep{zhou2025clip}. Although these approaches can significantly tighten verification for small perturbation radii, their effectiveness typically deteriorates as the perturbation radius increases due to the rapidly growing number of unstable neurons and increasingly loose LP relaxations. To bridge this gap, several works tighten LP-based verification by introducing multi-neuron or group-wise constraints that capture dependencies among multiple activations~\citep{singh2019beyond,muller2022prima,ferrari2022complete,ma2024relu}.

Recently, \citep{chiu2025sdp} integrated SDP relaxation into bound propagation, leading to significantly tighter $\ell_2$ verification with the scalability of bound propagation methods. In particular, \citep{chiu2025sdp} improves the relaxation of intermediate preactivations by propagating $\ell_2$-norm ball constraints through the network. However, its effectiveness can degrade significantly when the network is not Lipschitz-regularized, as illustrated in Table~\ref{tab:main_table}.

\subsection{Notations}
We use $\S^n$ to denote a set of symmetric $n\times n$ matrices. We use $\R^n_+$ and $\R^n_{++}$ to denote a set of nonnegative and positive length $n$ vectors. We use $x_i$ to denote the $i$-th entry of $x$. We use $\1$ to denote a column of ones. We use $\odot$ to denote elementwise multiplication. We use $e_i$ to denote $i$-th standard basis vector. We use $X\succeq 0$ to denote $X$ being positive semidefinite. We use $\diag(x)\in\S^n$ to denote a diagonal matrix whose diagonal is $x\in\R^n$ when applied to a vector, and use $\diag(X)\in\R^n$ to denote the diagonal entries of $X\in\S^n$ when applied to a matrix. We use $|\cdot|$ to denote the elementwise absolute value, and $\|\cdot\|_p$ to denote $\ell_p$ norm. We use $[\xlb,\xub]=\{x\mid \xlb_i\leq x_i\leq\xub_i\text{ for all $i$}\}$ to denote an elementwise interval between $\xlb$ and $\xub$.

\section{Background}

\subsection{Verification by bound propagation}

Given an input uncertainty set $\X_{0}$, the verification problem is to compute a provable lower bound on 
\[
f_{\star}=\min_{x^{(0)}\in\X_{0}}\inner{c^{(N)}}{x^{(N)}}\ \ \text{s.t.}\ \ x^{(k)}=\relu(z^{(k)}),\ \ z^{(k)}=W^{(k)}x^{(k-1)}+b^{(k)}\ \ \text{for $k=1,\ldots,N$}
\]
where $W^{(k)}$ and $b^{(k)}$ are the weights and biases, and $c^{(N)}$
is an output specification vector. We focus on the input set being an $\ell_p$-norm axis-aligned ellipsoid with center $\hat x$ and semiaxes $d>0$:
\[
\X_0=\E_p(\hat x,d)=\{\hat x + \diag(d)u\mid \|u\|_p\leq 1\}.
\]
We focus on $p=2$ in this work. Our method can be extended to any $1\leq p \leq \infty$, and we describe these extensions in Appendix~\ref{app:extension}. 
Bound propagation is effectively a dynamic programming algorithm for solving
this problem. At layer $k$, it constructs a lower bound
\[
\varphi_{k}(c)\le f_{k}(c)\coloneqq\min_{x\in\X_{k}}c^Tx,\qquad\text{where }\X_{k}\coloneqq\{\relu(W^{(k)}x+b^{(k)})\mid x\in\X_{k-1}\},
\]
assuming that a lower bound $\varphi_{k-1}(c)\le f_{k-1}(c)$ has already
been constructed for the previous layer. The base case is available in closed form
\[
\varphi_{0}(c)\coloneqq f_{0}(c)=\min_{x\in\X_{0}}c^Tx=c^T\hat x+\min_{\|u\|_p\leq 1}c^T\diag(d)u=c^T\hat x-\|c\odot d\|_q
\]
where $1/p+1/q=1$. By rolling out the induction, we obtain $\varphi_{N}(c^{(N)})\le f_{\star}$ at the end.

\subsection{$\alpha$-CROWN}
Given a valid lower bound $\varphi_{k-1}(\cdot)\le f_{k-1}(\cdot)$ for
the previous layer, $\alpha$-CROWN~\citep{xu2021fast} constructs a valid lower bound
$\varphi_{k}(c)\le f_{k}(c)$ in three steps.
The first step is forward propagation. Using $\varphi_{k-1}$, we obtain
interval bounds on the preactivation variable $z=W^{(k)}x+b^{(k)}$,
namely 
\[
\zlb_{i}\coloneqq\varphi_{k-1}(W^{(k)T}e_{i})+b_{i}^{(k)},\qquad\zub_{i}\coloneqq-\varphi_{k-1}(-W^{(k)T}e_{i})+b_{i}^{(k)},
\]
so that $z\in[\zlb,\zub]$. The $k$-th layer subproblem is augmented with redundant interval bounds as
\[
f_{k}(c)=\min_{x^{+}\in\X_{k}} c^Tx^{+}=\min_{\begin{subarray}{c}
z\in[\zlb,\zub]
\end{subarray}}\left\{c^T\relu(z):z=W^{(k)}x+b^{(k)},\ x\in\X_{k-1}\right\} .
\]
The second step is to replace the ReLU activation
by a linear lower bound. The key innovation of $\alpha$-CROWN is
to optimize over a family of lower bounds, parameterized by $\alpha\in[0,1]$,
such that 
\begin{equation}
g(\alpha)^Tz+h(\alpha)\leq c^T\relu(z)\quad\text{for all }z\in[\zlb,\zub].\label{eq:alpha-lb}
\end{equation}
The third step is back propagation. Substituting the linear lower
bound gives 
\begin{align*}
f_{k}(c) & \ge\min_{
z\in[\zlb,\zub]}\left\{g(\alpha)^Tz+h(\alpha):z=W^{(k)}x+b^{(k)},\ x\in\X_{k-1}\right\} \\
 & =\min_{x\in\X_{k-1}}\ip{g(\alpha)}{W^{(k)}x+b^{(k)}}+h(\alpha)= f_{k-1}(W^{(k)T}g(\alpha))+b^{(k)T}g(\alpha)+h(\alpha)\\
 & \ge\varphi_{k-1}(W^{(k)T}g(\alpha))+b^{(k)T}g(\alpha)+h(\alpha).
\end{align*}
Optimizing over all such lower bounds gives 
\[
\varphi_{k}(c)\coloneqq\max_{\alpha\in[0,1]}\;\varphi_{k-1}(W^{(k)T}g(\alpha))+b^{(k)T}g(\alpha)+h(\alpha).
\]
The validity $\varphi_{k}(c)\le f_{k}(c)$ follows from (\ref{eq:alpha-lb})
and the induction hypothesis $\varphi_{k-1}(c)\le f_{k-1}(c)$.

\subsection{Prior SDP-CROWN}

\global\long\def\B{\mathcal{B}}%
\global\long\def\I{\mathcal{I}}%
The prior SDP-CROWN~\citep{chiu2025sdp} extends $\alpha$-CROWN to work effectively for an axis-aligned $\ell_2$-norm ellipsoid input set. At each layer $k$, it assumes an outer ellipsoid approximation $\E_2(\hat{x},d_x)\supseteq\X_{k-1}$ for the previous layer. 
In the forward propagation step, it propagates the ellipsoid through the affine map, by defining a center $\hat{z}=W^{(k)}\hat{x}+b^{(k)}$ and semiaxes $d_z = \kappa a_z$ where
\begin{equation}\label{eq:ellipsoid_prop}
\kappa = \|\diag(a_z)^{-1}W^{(k)}\diag(d_x)\|_2,\quad a_z = \sqrt{(W^{(k)}\odot W^{(k)})(d_x \odot d_x)},
\end{equation}
and augmenting the $k$-th layer subproblem with both redundant interval and ellipsoid constraints. It then constructs a family of lower bounds, using the same
parameters $\alpha\in[0,1]$ as $\alpha$-CROWN, together with a new parameter $\lambda\ge0$ for the ellipsoid, such that 
\begin{equation}
g(\alpha)^Tz+h(\alpha,\lambda)\leq c^T\relu(z) \quad\text{for all }z\in[\zlb,\zub]\cap\E_2(\hat{z},d_z).\label{eq:sdp-lb}
\end{equation}
The key idea is to shrink the uncertainty set on which the lower bound must hold, thereby making the bound tighter. 
Finally, it back propagates the bound in the same way: 
\[
\varphi_{k}(c)\coloneqq\max_{\alpha\in[0,1],\lambda\ge0}\;\varphi_{k-1}(W^{(k)T}g(\alpha))+b^{(k)T}g(\alpha)+h(\alpha,\lambda).
\]
Then $\varphi_{k}(c)\le f_{k}(c)$ follows from (\ref{eq:sdp-lb}) and
the induction hypothesis $\varphi_{k-1}(c)\le f_{k-1}(c)$.

\section{Looseness in existing methods}

\global\long\def\Z{\mathcal{Z}}%
\global\long\def\Zout{\tilde{\Z}}%
\global\long\def\E{\mathcal{E}}%

\begin{figure}[t]
	\center
	\includegraphics[width=\textwidth]{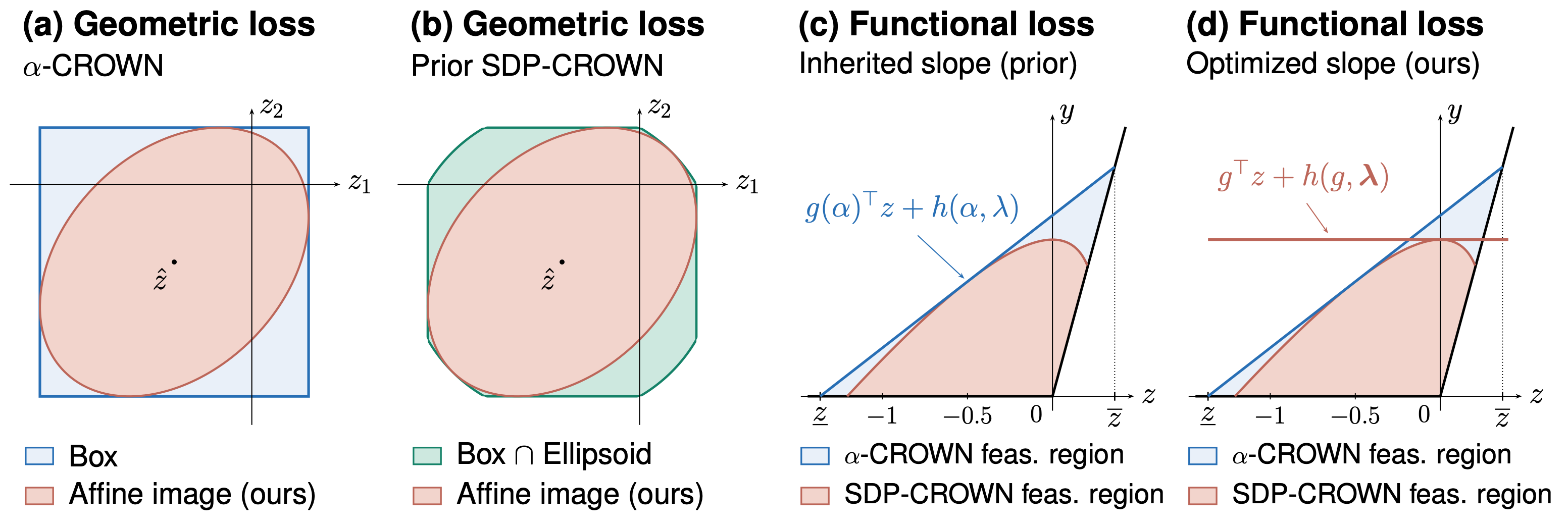}
	\caption{\textbf{Geometric and functional loss in bound propagation.} (a) $\alpha$-CROWN incurs geometric loss by enclosing the affine image in a box. (b) Prior SDP-CROWN reduces this loss by intersecting the box with an axis-aligned ellipsoid. Our method retains the affine image, avoiding this source of geometric loss. (c) In this example, the linear upper bound with inherited slope $g(\alpha)=\zub/(\zub-\zlb)$ cannot be tightened even though SDP-CROWN has a smaller feasible region. (d) We allow the slope $g$ to be optimized, allowing it to exploit the smaller feasible region.}\label{fig:main}
\end{figure}

The effectiveness of a verification method depends on the tightness of its certified lower bound $\varphi_N(c^{(N)})\le f_\star$. In bound propagation, looseness arises from two distinct sources.

\textbf{Geometric loss.} At the $k$-th layer, the preactivation set $\Z_{k}$ is relaxed to a tractable outer approximation 
\[
\Zout_{k}\supseteq\Z_{k}\coloneqq\{W^{(k)}x+b^{(k)}\mid x\in\X_{k-1}\}.
\]
The relevant question is: How much larger is $\Zout_{k}$ than the true set $\Z_{k}$?

\textbf{Functional loss.} The verifier then constructs affine lower
bounds on the ReLU objective over this tractable set: 
\[
g^Tz+h\le c^T\relu(z)\quad\text{for all }z\in\Zout_{k}.
\]
The relevant question is: How tightly can the nonlinear ReLU objective be lower bounded? 

\subsection{Looseness in $\alpha$-CROWN}
The bound propagation method $\alpha$-CROWN is designed to eliminate the functional loss. It uses an interval as the tractable outer approximation
\[
\Zout_{k}=[\zlb,\zub]\supseteq\Z_{k},
\]
so that the ReLU objective separates coordinatewise. This allows $\alpha$-CROWN
to construct an \emph{exactly tight} family of affine lower bounds,  $g(\alpha)^Tz+h(\alpha)$, in the sense that 
\[
\max_{\alpha\in[0,1]}\min_{z\in[\zlb,\zub]}\ g(\alpha)^Tz+h(\alpha) = \min_{z\in [\zlb,\zub]}\  c^T\relu(z).
\]
Specifically, for every $c$, there exists an $\alpha\in[0,1]$ that attains equality $c^T\relu(z)=g(\alpha)^Tz+h(\alpha)$ for some $z\in[\zlb,\zub]$ that minimizes $c^T\relu(z)$. Hence, $\alpha$-CROWN incurs no functional loss.

However, the tradeoff is potentially significant geometric loss. The
interval approximation $[\zlb,\zub]\supseteq\Z_{k}$ can dramatically increase the size of the preactivation set, particularly for $\ell_{2}$
adversaries. For example, if $\Z_{k}$ is an axis-aligned ellipsoid $\E_2(\hat z,d_z)$, then its elementwise interval is $\E_\infty(\hat z,d_z)=[\hat z-d_z,\hat z+d_z]$, whose Euclidean radius is larger by a factor of $\|d_z\|_2/\|d_z\|_\infty$
\[
\sup_{z\in\E_\infty(\hat z,d_z)}\|z-\hat z\|_{2}
=\frac{\|d_z\|_2}{\|d_z\|_\infty}\sup_{z\in\E_2(\hat z,d_z)}\|z-\hat z\|_{2}.
\]
Thus, replacing the ellipsoid by its coordinate box increases the maximum Euclidean deviation by a factor of $\|d_z\|_2/\|d_z\|_\infty$, reducing the tightness of $\alpha$-CROWN.

\subsection{Looseness in prior SDP-CROWN}
Prior SDP-CROWN reduces the geometric loss in $\alpha$-CROWN by choosing a more expressive tractable outer approximation, which is the intersection of an interval and an axis-aligned ellipsoid, 
\[
\Zout_{k}=[\zlb,\zub]\cap\E_2(\hat{z},d_z)\supseteq\Z_{k}.
\]
This directly targets the geometric loss caused by replacing an $\ell_{2}$-shaped uncertainty set by an interval. When $\Z_{k}$ is itself an axis-aligned ellipsoid $\E_2(\hat z,d_z)$, prior SDP-CROWN can represent the preactivation set exactly, hence yielding a factor of $\|d_z\|_2/\|d_z\|_\infty$ improvement over the interval relaxation $\E_\infty(\hat z,d_z)$ used by $\alpha$-CROWN. 

Unfortunately, this new tractable outer approximation is still not expressive
enough to be broadly useful for $\ell_{2}$ adversaries. In general, $\Z_{k}$ is an affine image under $\X_{k-1}$, for which an axis-aligned ellipsoid cannot closely capture. For instance, even when $\X_{k-1}$ is an axis-aligned ellipsoid $\E_2(\hat{x},d_x)$, its image under the affine transformation $z=W^{(k)}x+b^{(k)}$ is generally an ellipsoid that is not axis-aligned. The prior SDP-CROWN therefore encloses this ellipsoid in an axis-aligned ellipsoid $\E_2(\hat{z},d_z)$, with center $\hat z=W^{(k)}\hat x+b^{(k)}$ and semiaxes $d_z=\kappa a_z$ controlled by the spectral norm $\kappa=\|\diag(a_z)^{-1}W^{(k)}\diag(d_x)\|_{2}$ and $a_z$ defined in \eqref{eq:ellipsoid_prop}. This approximation can cause substantial enlargement along directions in which the original ellipsoid is narrow.

To better illustrate this point, write $W=\diag(d_z)^{-1}W^{(k)}\diag(d_x)$, and let $v_{\min}$ be a left singular vector associated with $\sigma_{\min}(W)$. Then the axis-aligned ellipsoid is larger than the true ellipsoid by a factor of $\cond(W)\coloneqq\sigma_{\max}(W)/\sigma_{\min}(W)$ in the direction $v_{\min}$: 
\begin{align*}
&\sup_{z\in\E_2(\hat z,d_z)}\left|\ip{v_{\min}}{\diag(d_z)^{-1}(z-\hat z)}\right|
=\sup_{\|u\|_2\leq 1}\left|\ip{v_{\min}}{u}\right|
=\cond(W)\sup_{\|u\|_2\leq 1}\left|\ip{v_{\min}}{Wu}\right|\\
&=\cond(W)\sup_{z}\left\{\left|\ip{v_{\min}}{\diag(d_z)^{-1}(z-\hat z)}\right|:z = W^{(k)}x+b^{(k)},\ x\in\E_2(\hat{x},d_x) \right\}.
\end{align*}
The enclosing ellipsoid is therefore $\cond(W)$ times wider in this normalized direction. Hence, poor conditioning $\cond(W)\to\infty$ can cause substantial enlargement along directions in which the affine image is narrow, reducing the tightness of prior SDP-CROWN. We extend the prior SDP-CROWN to construct the outer approximation using an affine image, thereby reducing the geometric loss.

In addition to geometric loss, the prior SDP-CROWN also incurs functional loss in the sense that
\[
\max_{\alpha\in[0,1],\lambda\geq 0}\min_{z\in[\zlb,\zub]\cap\E_2(\hat z,d_z)}\ g(\alpha)^Tz+h(\alpha,\lambda) \leq  \min_{z\in[\zlb,\zub]\cap\E_2(\hat z,d_z)}\  c^T\relu(z).
\]
For a zero-centered ellipsoid, i.e., $\hat z=0$ and $-\zlb=\zub=d_z$, \cite{chiu2025sdp} show that the bound used by the prior SDP-CROWN is exactly tight. Specifically, for every $c$, there exists an $\alpha\in[0,1]$ and $\lambda\geq 0$ that attains equality $c^T\relu(z)=g(\alpha)^Tz+h(\alpha,\lambda)$ for some $z\in[-d_z,d_z]\cap\E_2(0,d_z)$ that minimizes $c^T\relu(z)$. However, when the ellipsoid is not zero-centered, the prior SDP-CROWN can suffer from functional loss. The central weakness in the prior SDP-CROWN is to inherit the slope $g = g(\alpha)$ from $\alpha$-CROWN, which limits how well the linear bound exploits the ellipsoidal domain even after optimizing $\alpha$ and $\lambda$, introducing functional loss as shown in Figure~\ref{fig:main}. We jointly optimize $g$ and $h(g,\blambda)$ over domains strengthened by affine-image constraints, which substantially improve verification performance in our experiments.

\section{This work: SDP-CROWN with affine-image propagation}\label{sec:method}

\global\long\def\Xout{\tilde{\X}}%
\global\long\def\xlb{\underline{x}}%
\global\long\def\xub{\overline{x}}%
In this paper, we improve the prior SDP-CROWN by augmenting ellipsoid constraints with \emph{affine-image constraints}. Instead of approximating the preactivation set only by an intersection of  an interval and an axis-aligned ellipsoid $[\zlb,\zub]\cap \E_2(\hat{z},d_z)$, as in prior SDP-CROWN, we additionally intersect with an affine image of an $\ell_p$-norm ellipsoid:
\begin{align}\label{eq:our_set}
\Zout_{k} & =[\zlb,\zub]\cap \E_2(\hat{z},d_z) \cap\{W^{(k)}x+b^{(k)}\mid x\in\E_p(\hat{x},d_x)\}\quad\text{with}\quad \hat z = W^{(k)}\hat x+b^{(k)}.
\end{align}
This seemingly small change in representation directly tackles the
geometric loss identified in the previous section. The affine-image (\ref{eq:our_set}) also allows our method to naturally handle verification problems with general $\ell_p$-norm input sets $\X_0$. In contrast, the ellipsoid propagation in prior SDP-CROWN is limited to $\ell_2$-norm input sets. We focus on the $\ell_2$ case, i.e., $p=2$ in  (\ref{eq:our_set}), in the main paper and defer the $\ell_p$-norm generalization to  Appendix~\ref{app:extension}.
 
At the input layer, the improvement is immediate. If the input uncertainty
set $\X_{0}=\E_2(\hat x,d)$ is an ellipsoid, then the first preactivation set is exactly the affine image of that ellipsoid. Hence the representation (\ref{eq:our_set}) gives $\Zout_{1}=\Z_{1}$ and incurs no geometric loss at the input. This significantly improves upon $\alpha$-CROWN and prior SDP-CROWN respectively, as previously described. 

But the same representation is also much more useful in the intermediate layers.
To make the accounting precise, let $\Xout_{k-1}\supseteq\X_{k-1}$
denote the outer approximation carried from the previous layer. Our relaxation uses 
\[
\Zout_{k}=[\zlb,\zub]\cap \E_2(\hat{z},d_z)\cap\{W^{(k)}x+b^{(k)}\mid x\in\Xout_{k-1}\}
\supseteq 
\{W^{(k)}x+b^{(k)}\mid x\in\X_{k-1}\}=\Z_{k}
\]
to approximate the exact preactivation set 
$\Z_{k}$, where the affine map $W^{(k)}x+b^{(k)}$ itself is applied exactly; if $\Xout_{k-1}=\X_{k-1}$, then $\Zout_{k}=\Z_{k}$, up to the redundant valid interval constraints.
More generally, the only geometric loss is the loss already present
in the previous-layer relaxation $\Xout_{k-1}\supseteq\X_{k-1}$.
This is the key distinction from prior methods, for which only our representation maintains the affine geometry and cross-neuron correlations encoded by $W^{(k)}$.

The remaining challenge is to keep functional loss small. Following the same
semidefinite programming steps as \citep{chiu2025sdp} to derive a lower bound on the affine image $\{Wx+b\mid x\in\E_2(\hat{x},d_x)\}$ leads to semidefinite constraints of the form \begin{equation}
\sigma I \succeq \diag(d_x)W^{T}\diag(\lambda)W\diag(d_x),\label{eq:tau-WtlamW}
\end{equation}
where $\lambda\ge0$ and $\sigma\ge0$ are Lagrange multipliers that
must be optimized as part of the bound. Enforcing this constraint
exactly would indeed yield very small functional loss, but would also
require solving a large and fully dense SDP at every step. That is far
too expensive for a scalable verifier.

The prior SDP-CROWN was able to work on a large scale with bound propagation because it replaced (\ref{eq:tau-WtlamW}) with the spectral sufficient condition 
\begin{equation}
\sigma \geq \|W\diag(d_x)\|_{2}^{2}\|\lambda\|_\infty.\label{eq:spec_suff}
\end{equation}
This crude sufficient condition (\ref{eq:spec_suff}) is sufficiently tight when $W$ has small Lipschitz constant, which is why SDP-CROWN was able to achieve very small functional loss in Lipschitz-regularized networks. For more general ellipsoids, our key observation is that one can keep the affine image while replacing the full SDP constraint by a scaled diagonal-dominance surrogate.
For example, a scaled diagonal-dominance condition takes the form
\begin{align*}
\sigma &\geq\|\gamma(\diag(d_x)W^{T}\diag(\lambda)W\diag(d_x),l)\|_\infty\quad\text{for some $l>0$}
\end{align*}
where $\gamma: \S^n\times \R^{n}_{++}\to \R^{n}_{+}$ is the scaled Gershgorin operation
\begin{equation}\label{eq:gershgorin_operation}
\gamma_i(M,l) = \max\left\{0,M_{ii}+\sum_{j\neq i}\frac{l_j}{l_i}|M_{ij}|\right\}\quad \text{for $i=1,\ldots,n$}
\end{equation}
We use scaled diagonal-dominance to obtain tractable sufficient conditions while retaining the affine-image constraints. It preserves the directions
and correlations encoded by $W$, while remaining compatible with the back-propagation structure of the bound propagation framework. This approach can be particularly effective for convolutional layers, for which $W^TW$ could naturally exhibit diagonal-dominant structure, as illustrate in Figure~\ref{fig:dd}. The resulting lower bound is stated below.
\begin{theorem}
\label{thm:sdp_crown} Given $c\in\R^{m}$, an affine-image $\{Wx+b\mid x\in\E_2(\hat{x},d_x)\}$ with $W\in\R^{m\times n}$, an ellipsoid $\E_2(\hat{z},d_z)$ with $\hat{z}=W\hat{x}+b$, and an interval $[\zlb,\zub]$ with $\zlb,\zub\in\R^m$. The following holds 
\[
g^{T}z+h(g,\blambda)\leq c^{T}\relu(z)\quad\text{for all } z\in [\zlb,\zub]\cap\E_2(\hat{z},d_z)\cap\{Wx+b\mid x\in\E_2(\hat{x},d_x)\}
\]
for any $g\in\R^m$, $\blambda=(\lambda,\tau,l,\eta)\in\R^{m}_+\times \R^{m}_+\times \R^{n}_{++}\times \R_+$, and 
\[
h(g,\blambda)=-\tau^T\qub-\frac{1}{2}\left[\eta+\|\gamma(M(\lambda,\eta),l)\|_{\infty}+\sum_{j=1}^{m}\left(\frac{\phi_{j}^{2}}{\lambda_{j}}-\lambda_{j}\hat z_{j}^2\right)\right]
\]
where $\gamma$ is the scaled Gershgorin operation defined in (\ref{eq:gershgorin_operation}), and
\begin{gather*}
M(\lambda,\eta)=\diag(d_x)W^T\left[\diag(\lambda)-\eta\diag(d_z)^{-2}\right]W\diag(d_x),\\
	\phi_{j} =\min\left\{c_{j}-g_{j}-\tau_{j}(\pub_{j}-1)-\lambda_{j}\hat{z}_{j},\ g_{j}+\tau_{j}\pub_{j}+\lambda_{j}\hat{z}_{j},\ 0\right\}
\end{gather*}
for $j=1,\ldots,m$. Here, $\pub_j,\qub_j$ are the parameters of the linear upper bound $\relu(z_j)\leq \pub_jz_j+\qub_j$ for all $z_j\in[\zlb_j,\zub_j]$, and are  defined in (\ref{eq:beta_gamma}).
\end{theorem}

\subsection{Integration with bound propagation}
Theorem~\ref{thm:sdp_crown} can be inserted into bound propagation. At layer $k$, we first propagate the interval bounds $[\zlb,\zub]$ and the affine-image representation to construct $\Zout_{k}$.
Then, for a backward specification vector $c$, Theorem~\ref{thm:sdp_crown} gives a valid affine lower bound on $c^{T}\relu(z)$ over $\Zout_{k}$. Substituting $z=W^{(k)}x+b^{(k)}$ gives the recursive bound 
\[
f_{k}(c)\ge f_{k-1}(W^{(k)T}g)+g^{T}b^{(k)}+h(g,\blambda).
\]
Thus, the propagated lower bound is 
\[
\varphi_{k}(c)=\max_{g,\blambda}\left\{ \varphi_{k-1}(W^{(k)T}g)+g^{T}b^{(k)}+h(g,\blambda)\right\} ,
\]
where the maximization is over the admissible parameters $g$ and $\blambda$ in Theorem~\ref{thm:sdp_crown}. We note that we allow $g$ to be optimized freely in this work, as opposed to prior SDP-CROWN, which inherits $g=g(\alpha)$ from $\alpha$-CROWN. The backward pass sends $W^{(k)T}g$ to the previous layer and accumulates the scalar offset $b^{(k)T}g+h(g,\blambda)$. Since the parameters $g,\blambda$ are either unconstrained or nonnegative/positive, they can be optimized by projected gradient ascent similar to other bound propagation methods.

The only major additional computation is the diagonal-dominance surrogate $\|\gamma(M(\lambda,\eta),l)\|_\infty$. To illustrate the computation, consider $d_x=\1$, $d_z=\1$, $l=\1$ and $\eta = 0$ as they do not change the structure in $W$. When $W$
corresponds to a convolutional layer, the sparse convolutional structure is preserved during the computation of $\|\gamma(M(\lambda,0),\1)\|_\infty=\|W^T \diag(\lambda) W\|_\infty$. In particular, two input neurons can have a nonzero interaction in $W^T\diag(\lambda) W$ only if they appear together in at least one convolution window. If $W$ has kernel size
$k_h \times k_w$, then the resulting interaction pattern has effective
spatial support at most $(2k_h-1)\times(2k_w-1)$. Our GPU-optimized implementation exploits this sparsity directly to construct the matrix $W^T\diag(\lambda) W$, by enumerating the local interactions generated from each convolution window in parallel and accumulating only the
nonzero terms.

As a result, our method preserves the computational structure of bound propagation while retaining more of the geometry of preactivation sets. Compared with prior SDP-CROWN, it retains affine-image constraints alongside axis-aligned enclosures, reducing geometric loss. Compared with full SDP verifiers, it avoids dense semidefinite optimization. This offers a practical balance between tightness and computational cost while retaining the recursive, GPU-accelerated backward pass of existing bound propagation methods.

\subsection{Integration with branch-and-bound}
\input{sections/bab}

\section{Experiments}
\input{sections/experiment}


\section*{Acknowledgements and funding disclosures} 
Financial support for this work was provided by NSF CAREER Award ECCS-2047462, and ONR Award N00014-24-1-2671. Huan Zhang is supported in part by the AI2050 program at Schmidt Sciences (AI2050 Early Career Fellowship) and NSF (2525287, 2331967).

\newpage

\bibliography{robustness}

\newpage
\appendix

\section{Proof of Theorem~\ref{thm:sdp_crown}}
We start by introducing our absolute-split SDP relaxation and then provide the proof of our main theorem in this section.

\begin{figure}[t!]
    \centering
    \includegraphics[width=0.33\textwidth]{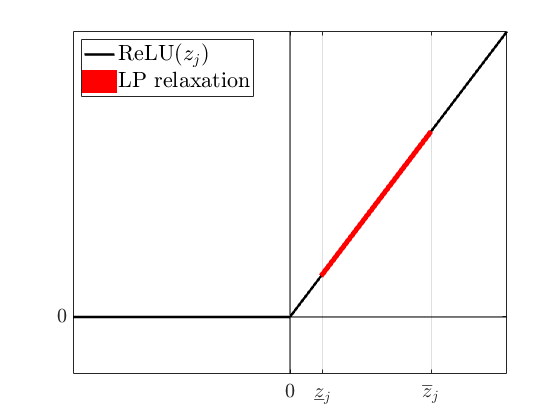}%
    \includegraphics[width=0.33\textwidth]{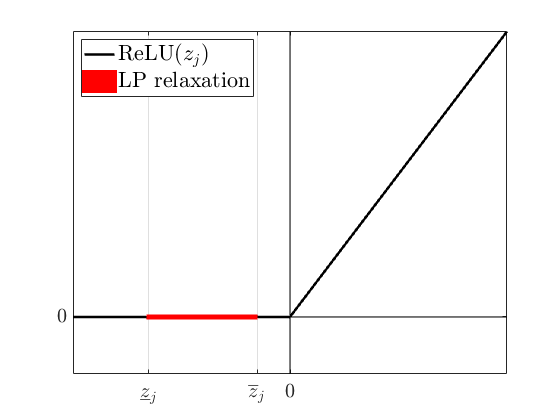}%
    \includegraphics[width=0.33\textwidth]{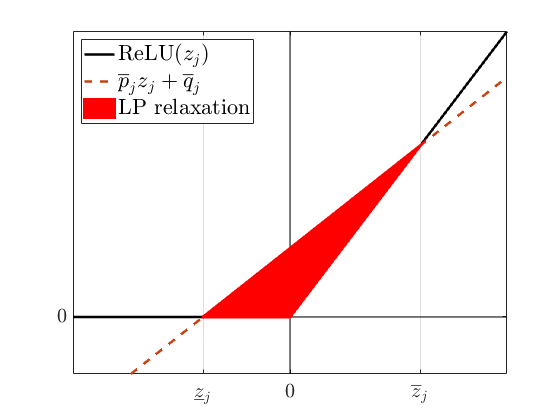}
    \includegraphics[width=0.33\textwidth]{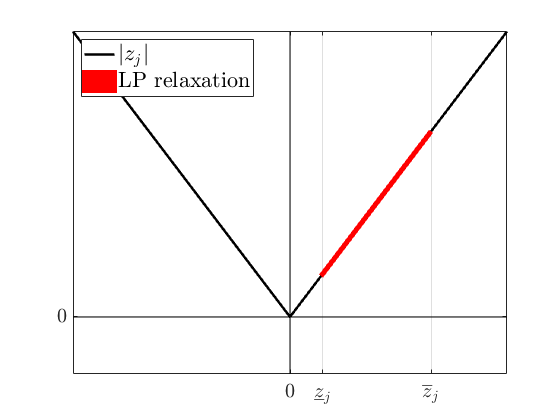}%
    \includegraphics[width=0.33\textwidth]{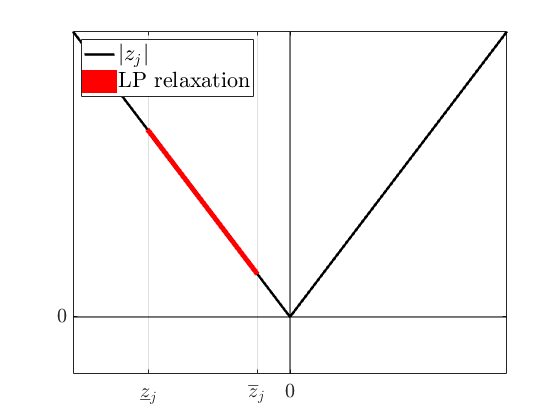}%
    \includegraphics[width=0.33\textwidth]{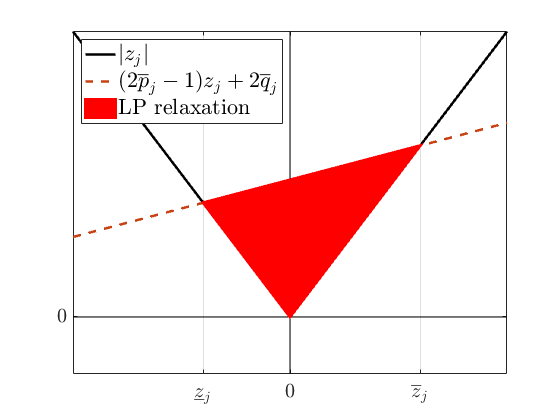}%
    \caption{\textbf{Comparison of the triangular relaxation \citep{ehlers2017formal} and the absolute-split relaxation \eqref{eq:lp_relu_abs}.} \textbf{(Left.)} $\underline z_j \geq 0$. In this case, $y_j=z_j$ and $r_j=z_j$. \textbf{(Middle.)} $\overline z_j\leq 0$. In this case, $y_j=0$ while $r_j=-z_j$. \textbf{(Right.)} $\underline z_j\leq 0\leq \overline z_j$. In this case, $\relu(z_j)\leq y_j\leq \pub_j z_j+\qub_j$ while $|z_j|\leq r_j\leq (2\pub_j-1) z_j+2\qub_j$.}\label{fig:abs_split}
\end{figure}
\subsection{The absolute-split SDP relaxation}
Our derivation of SDP relaxations for $y=\relu(z)$ is based on the absolute-split $y=\frac{1}{2}(z+|z|)$. Through this view, it naturally exposes the structure in \eqref{eq:tau-WtlamW}. Given $\underline z\leq z\leq \overline z$ where $z\in\R^m$, we start by developing a linear relaxation by characterizing the convex hull of $r=|z|$:
\begin{equation}\label{eq:lp_relu_abs}
\begin{aligned}
y_j = \frac{1}{2}(z_j+r_j),\quad |z_j|\leq r_j\leq (2\pub_j-1)z_j+2\qub_j\quad\text{for $j=1,\ldots,m$}
\end{aligned}
\end{equation}
\begin{equation}\label{eq:beta_gamma}
\text{where}\quad \pub_j =
\begin{cases}
1 & \text{if } \underline{z}_j \geq 0,\\
0 & \text{if } \overline{z}_j \leq 0,\\
\frac{\overline{z}_j}{\overline{z}_j - \underline{z}_j} & \text{otherwise,}
\end{cases}
\quad
\qub_j =
\begin{cases}
0 & \text{if } \underline{z}_j \geq 0,\\
0 & \text{if } \overline{z}_j \leq 0,\\
-\frac{\overline{z}_j \, \underline{z}_j}{\overline{z}_j - \underline{z}_j} & \text{otherwise.}
\end{cases}
\end{equation}
We note that the linear relaxation derived from the absolute-split (\ref{eq:lp_relu_abs}) is equivalent to the triangular relaxation $\relu(z_j)\leq y_j\leq \pub_j z_j+\qub_j$ of \citep{ehlers2017formal}. We provide an illustration in Figure~\ref{fig:abs_split}. The absolute-split SDP relaxation can be derived based on the  observation that \eqref{eq:lp_relu_abs} becomes \emph{exact} after adding a nonlinear inequality constraint $y=\relu(z)\ \iff\ \eqref{eq:lp_relu_abs},\quad r\odot r \leq z\odot z$.\footnote{Though the linear bound $r\leq (2\pub-1)\odot z+2\qub$ becomes redundant after adding $r\odot r \leq z\odot z$, it will ensure the subsequent absolute-split SDP relaxation will always be at least as tight, and potentially strictly tighter than the absolute-split LP relaxation \eqref{eq:lp_relu_abs}. This is commonly refer as the strengthening constraints for SDP relaxations \citep{brown2022unified}}

Given a weight $W\in\R^{m\times n}$ and bias $b\in\R^m$. Substituting $z=Wx+b$ and relaxing $xx^T\preceq X$ and $r_j\cdot r_j \leq R_j$ yield the following SDP relaxation
\begin{equation}\label{eq:sdp_relu_abs}
\begin{gathered}
\eqref{eq:lp_relu_abs},\quad
R \leq \diag(WXW^T)+2z\odot b - b\odot b,\quad z=Wx+b,\\ 
 \begin{bmatrix}1&x^T\\x&X\end{bmatrix}\succeq 0,\quad \begin{bmatrix}1&r_j\\r_j&R_j\end{bmatrix}\succeq 0\quad\text{for all $j=1,\ldots,m$.}
\end{gathered}
\end{equation}

\subsection{Proof of Theorem~\ref{thm:sdp_crown}}
Theorem~\ref{thm:sdp_crown} can be proved by deriving the SDP relaxation of the following
\begin{equation*}
\min_{z\in\R^m}\ c^T\relu(z)\quad\text{s.t.}\quad z\in [\zlb,\zub]\cap\E_2(\hat{z},d_z)\cap\{Wx+b\mid x\in\E_2(\hat{x},d_x)\}.
\end{equation*}
We start by applying a variable substitution $x\equiv x+\hat x$ and $\hat z = W\hat x + b$ so that the constraints on $z$ become
\begin{equation*}
z\in [\zlb,\zub]\cap\E_2(\hat{z},d_z)\cap\{Wx+\hat z\mid x\in\E_2(0,d_x)\}
\end{equation*}
and then applying the absolute split SDP relaxation \eqref{eq:sdp_relu_abs} to yield
\begin{equation}\label{eq:thm1_primal}
\min_{\substack{x\in\R^n,X\in\S^n\\
z,r,R\in\R^m}}\frac{1}{2}c^T(z+r)\quad \text{s.t.}\quad
\begin{array}{c}
	|z|\leq r \leq (2\pub-1)\odot z + 2\qub, \quad z = Wx+\hat z,\\
R \leq \diag(WXW^T) + 2z\odot \hat z - \hat z\odot \hat z,\\
\tr(\diag(d_x)^{-2}X)\leq 1,\quad \tr(\diag(d_z)^{-2}WXW^T)\leq 1,\\
 \begin{bmatrix}
	1 & x^T \\ x & X
\end{bmatrix}\succeq 0,\quad \begin{bmatrix}
	1 & r_j \\ r_j & R_j
\end{bmatrix}\succeq 0\quad\text{for $j=1,\ldots,m$}
\end{array}
\end{equation}
Notice that the SDP relaxations for $x\in\E_2(0,d_x)\iff \tr(\diag(d_x)^{-1}xx^T\diag(d_x)^{-1})\leq 1$ and $z\in \E_2(\hat{z},d_z)\iff \tr(\diag(d_z)^{-1}Wxx^TW^T\diag(d_z)^{-1})\leq 1$ are obtained by relaxing $xx^T\preceq X$. The core insight to prove Theorem~\ref{thm:sdp_crown} is to fix any $x\in\E_2(0,d_x)$ and $z=Wx+\hat z\in\Zout$ where $\Zout$ is defined in \eqref{eq:our_set} with $p=2$,  then take the dual to lower bound the above. In particular, the Lagrangian dual is given by:
\begin{equation}\label{eq:ellipsoid_dual}
 \begin{aligned}
	\max_{\substack{\mu,s,t,\tau,\lambda\in\R^m_+ \\\sigma,\eta\in\R_+}}\ & g^Tz-2\tau^T\qub+\frac{1}{2}x^TMx-\frac{1}{2}\left(\sigma+\eta-\hat z^T\diag(\lambda)\hat z+\mu^T\1\right)\\
	\text{s.t. }\quad\ & g = \frac{1}{2}c + s - t - (2\pub-1)\odot\tau - \lambda\odot\hat z, \quad \phi = \frac{1}{2}c-s-t+\tau,\\
	& M = \sigma\diag(d_x)^{-2}- W^T\left[\diag(\lambda)-\eta\diag(d_z)^{-2}\right]W\succeq 0,\\
	&\mu_j\lambda_j\geq \phi_j^2\quad\text{for $j=1,\ldots,m$}.
 \end{aligned}
\end{equation}
We defer the detailed derivation for the dual problem (\ref{eq:ellipsoid_dual}) to Appendix~\ref{app:dual}. The dual variable $M$ is fully dense, and therefore $M\succeq 0$ cannot be enforced efficiently. We propose to instead optimize over the scaled diagonal-dominance surrogate for $M$:
\[
	l > 0,\quad \sigma\1 \geq \gamma\Big(\underbrace{\diag(d_x)W^T\left[\diag(\lambda)-\eta\diag(d_z)^{-2}\right]W\diag(d_x)}_{M(\lambda,\eta)},l\Big) \implies M\succeq 0
\]
where $\gamma$ is the scaled Gershgorin operation defined in (\ref{eq:gershgorin_operation}). Since setting $\sigma = \|\gamma(M(\lambda,\eta),l)\|_\infty$ ensures $M\succeq 0$, (\ref{eq:ellipsoid_dual}) can be relaxed by setting $\sigma = \|\gamma(M(\lambda,\eta),l)\|_\infty$  and observing that
\begin{align*}
\frac{1}{2}(x^TMx-\sigma)&=\frac{1}{2}\left(\diag(d_x)^{-1}x\right)^T\left(\sigma I - M(\lambda,\eta)\right)\left(\diag(d_x)^{-1}x\right)-\frac{1}{2}\sigma\\
&\geq -\frac{1}{2} \|\gamma(M(\lambda,\eta),l)\|_\infty.
\end{align*}
Hence, the diagonal-dominant surrogate yields the following relaxation for \eqref{eq:ellipsoid_dual}
\begin{equation*}
 	\begin{aligned}
 		\max_{\substack{\mu,s,t,\tau,\lambda\in\R^m_+ \\l\in\R^n_{++},\eta\in\R_+}}\ & g^Tz-2\tau^T\qub-\frac{1}{2}\left(\eta+\|\gamma(M(\lambda,\eta),l)\|_\infty-\hat z^T\diag(\lambda)\hat z+\mu^T\1\right)\\
	\text{s.t. }\quad\ & g = \frac{1}{2}c + s - t - (2\pub-1)\odot\tau - \lambda\odot\hat z, \quad \phi = \frac{1}{2}c-s-t+\tau,\\
  		&\mu_j\lambda_j \geq  \phi_j^2\quad\text{for $j=1,\ldots,m$}
 	\end{aligned}
\end{equation*}
The variables $s,t\geq 0$ can be eliminated by applying the following variable transformation
\begin{equation}\label{eq:st}
 s\geq 0,\quad t\geq 0\quad\iff\quad \theta=s+t,\quad \omega = s-t,\quad \theta\geq |\omega|.
\end{equation}
Notice that each $\theta_j$ in $\phi_j$ can be optimized in closed form as 
\[
\min_{\theta_j\geq |\omega_j|}\phi_j^2=\min\{c_j/2-|\omega_j|+\tau_j,\ 0\}^2 = \min\{c_j/2-\omega_j+\tau_j,\ c_j/2+\omega_j+\tau_j,\ 0\}^2
\]
and $\omega_j$ can be eliminated by substituting $g_j = c_j/2 + \omega_j - \tau_j(2\pub_j-1) - \lambda_j\hat z_j$ into $\phi_j$ to yield
\[
\phi_j = \min\{c_j-g_j-2\tau_j(\pub_j-1)-\lambda_j\hat z_j,\ g_j+2\tau_j\pub_j+\lambda_j\hat z_j,\ 0\}.
\]
Finally, applying the above two variable substitutions yields
\begin{equation*}
 	\begin{aligned}
 		\max_{\substack{\mu,\tau,\lambda\in\R^m_+,g\in\R^m\\
 		l\in\R^n_{++},\eta\in\R_+}}\ & g^Tz-2\tau^T\qub-\frac{1}{2}\left(\eta+\|\gamma(M(\lambda,\eta),l)\|_\infty-\hat z^T\diag(\lambda)\hat z+\mu^T\1\right)\\
 		\text{s.t. }\quad\ & \phi_j = \min\{c_j-g_j-2\tau_j(\pub_j-1)-\lambda_j\hat z_j,\ g_j+2\tau_j\pub_j+\lambda_j\hat z_j,\ 0\},\quad \mu_j\lambda_j\geq \phi_j^2
 	\end{aligned}
\end{equation*}
for $j=1,\ldots,m$. We are now ready to prove Theorem~\ref{thm:sdp_crown}.
\begin{proof}
	Rescaling $\tau\equiv\frac{1}{2}\tau$, the desired result in  Theorem~\ref{thm:sdp_crown} follows from fixing any $\blambda=(\lambda,\tau,l,\eta)\in\R^{m}_+\times \R^{m}_+\times \R^{n}_{++}\times \R_+$ and optimizing over $\mu$
\begin{equation*}
 	\begin{aligned}
 		\max_{\substack{g,\blambda}}&\ g^Tz+h(g,\blambda)\\
 		\text{s.t. }\  & h(g,\blambda)=-\tau^T\qub-\frac{1}{2}\cdot\left(\eta+\|\gamma(M(\lambda,\eta),l)\|_\infty+\sum_{j=1}^{m}\frac{\phi_{j}^{2}}{\lambda_{j}}-\lambda_{j}\hat z_{j}^2\right)\\
 		& \phi_j = \min\{c_j-g_j-\tau_j(\pub_j-1)-\lambda_j\hat z_j,\ g_j+\tau_j\pub_j+\lambda_j\hat z_j,\ 0\}.
 	\end{aligned}
 \end{equation*}
for $j=1,\dots,m$. Since for every fixed $z\in\Zout$ the above is a lower bound of $c^T\relu(z)$, it then follows that $g^Tz+h(g,\blambda)\leq c^T\relu(z)$ for all $z\in\Zout$.
\end{proof}

\section{Extension for our method}\label{app:extension}
In this section, we show that Theorem~\ref{thm:sdp_crown} can be extended to handle an affine image under $\ell_p$-norm ellipsoid
\[
\{Wx+b\mid x\in\E_p(\hat{x},d_x)\}
\]
for any $1\leq p\leq \infty$. We note that when $p=\infty$, the $\ell_\infty$-norm ellipsoid is equivalent to an elementwise interval $\E_\infty(\hat x,d_x)=[\hat x-d_x,\hat x+d_x]$, and therefore it automatically captures the elementwise input set. In the following corollary, we develop a linear lower bound on (\ref{eq:our_set}) for $1\leq p\leq\infty$.

\begin{corollary}\label{cor:sdp_crown} 
Given $c\in\R^{m}$, an affine-image $\{Wx+b\mid x\in\E_p(\hat{x},d_x)\}$ with $W\in\R^{m\times n}$ and $1\leq p\leq \infty$, an ellipsoid $\E_2(\hat{z},d_z)$ with $\hat{z}=W\hat{x}+b$, and an interval $[\zlb,\zub]$ with $\zlb,\zub\in\R^m$. The following holds 
\[
g^{T}z+h(g,\blambda)\leq c^{T}\relu(z)\quad\text{for all } z\in [\zlb,\zub]\cap\E_2(\hat{z},d_z)\cap\{Wx+b\mid x\in\E_p(\hat{x},d_x)\}
\]
for any $g\in\R^m$, $\blambda=(\lambda,\tau,l,\eta)\in\R^{m}_+\times \R^{m}_+\times \R^{n}_{++}\times \R_+$, and 
\[
h(g,\blambda)=-\tau^T\qub-\frac{1}{2}\left[\eta+\|\gamma(M(\lambda,\eta),l)\|_{q}+\sum_{j=1}^{m}\left(\frac{\phi_{j}^{2}}{\lambda_{j}}-\lambda_{j}\hat z_{j}^2\right)\right],\quad \frac{1}{\max\{1,p/2\}}+\frac{1}{q}= 1
\]
where $\gamma$ is the scaled Gershgorin operation defined in (\ref{eq:gershgorin_operation}), $\qub$ is defined in (\ref{eq:beta_gamma}), $M(\lambda,\eta)$ and $\phi$ are defined in Theorem~\ref{thm:sdp_crown}.
\end{corollary}

\subsection{Proof of Corollary~\ref{cor:sdp_crown}}
The proof of Corollary~\ref{cor:sdp_crown} follows the same structure as the proof of Theorem~\ref{thm:sdp_crown}. We start by  deriving the SDP relaxation of the following
\begin{equation*}
\min_{z\in\R^m}\ c^T\relu(z)\quad\text{s.t.}\quad z\in [\zlb,\zub]\cap\E_2(\hat{z},d_z)\cap\{Wx+b\mid x\in\E_p(\hat{x},d_x)\}.
\end{equation*} 
The SDP relaxation of the above is given by replacing the constraint $\tr(\diag(d_x)^{-1}X\diag(d_x)^{-1})\leq 1$ in the primal problem (\ref{eq:thm1_primal}) with
\[
\begin{cases}
	\|\diag\left(\diag(d_x)^{-1}X\diag(d_x)^{-1}\right)\|_{p/2}\leq 1 & \text{if $2\leq p\leq \infty$}\\
	\|\diag\left(\diag(d_x)^{-1}X\diag(d_x)^{-1}\right)\|_{1}\leq 1 & \text{if $1\leq p < 2$}.
\end{cases}
\]
To maintain convexity, we use the $\ell_1$ unit ball when $1\leq p < 2$, which is the convex hull of $\ell_{p/2}$ quasi-norm unit ball  for $1\leq p < 2$. The proof of Corollary~\ref{cor:sdp_crown} follows from fixing any $x\in\E_p(0,d_x)$ and $z=Wx+\hat z\in\Zout$ where $\Zout$ is defined in \eqref{eq:our_set}, and taking the dual to lower bound the primal problem. To see how its dual differ from  (\ref{eq:ellipsoid_dual}), let $\sigma\in\R^n$ be the dual variable of $v=\diag\left(\diag(d_x)^{-1}X\diag(d_x)^{-1}\right)$, the Lagrangian is
\[
\inner{\sigma }{\diag\left(\diag(d_x)^{-1}X\diag(d_x)^{-1}\right)-v}=\inner{X}{\diag(d_x)^{-1}\diag(\sigma)\diag(d_x)^{-1}}-\sigma^Tv.
\]
Following the same derivation in Appendix~\ref{app:dual}, the first term in the Lagrangian contributes to the PSD constraints
\[
\diag(\sigma)\diag(d_x)^{-2}- W^T\left[\diag(\lambda)-\eta\diag(d_z)^{-2}\right]W\succeq 0,
\]
where $\eta$ is the dual variable associated with the constraint $\tr(\diag(d_z)^{-2}WXW^T)\leq 1$ in (\ref{eq:thm1_primal}).

For the second term $-\sigma^Tv$ in the Lagrangian, notice that $\|v\|_{\max\{1,p/2\}}\leq 1$, taking the infimum over $\|v\|_{\max\{1,p/2\}}\leq 1$ yields
\[
\inf_{\|v\|_{\max\{1,p/2\}}\leq 1} -\sigma^Tv = -\|\sigma\|_q,\quad \text{where}\quad \frac{1}{\max\{1,p/2\}}+\frac{1}{q}= 1
\]
Rescaling $\sigma\equiv \frac{1}{2}\sigma$ and $\eta\equiv \frac{1}{2}\eta$, the Lagrangian dual is given by:
\begin{equation}\label{eq:general_dual}
 \begin{aligned}
	\max_{\substack{\mu,s,t,\tau,\lambda\in\R^m_+ \\\sigma\in\R^n,\eta\in\R_+}}\ & g^Tz-2\tau^T\qub+\frac{1}{2}x^TMx-\frac{1}{2}\left(\|\sigma\|_q+\eta-\hat z^T\diag(\lambda)\hat z+\mu^T\1\right)\\
	\text{s.t. }\quad\ & g = \frac{1}{2}c + s - t - (2\pub-1)\odot\tau - \lambda\odot\hat z, \quad \phi = \frac{1}{2}c-s-t+\tau,\\
	& M = \diag(\sigma)\diag(d_x)^{-2}- W^T\left[\diag(\lambda)-\eta\diag(d_z)^{-2}\right]W\succeq 0,\\
	&\mu_j\lambda_j\geq \phi_j^2\quad\text{for $j=1,\ldots,m$}.
 \end{aligned}
\end{equation}
Again, the dual variable $M$ is fully dense, and therefore $M\succeq 0$ cannot be enforced efficiently. We propose to instead optimize over the scaled diagonal-dominance surrogate for $M$
\[
	l > 0,\quad \sigma \geq \gamma\Big(\underbrace{\diag(d_x)W^T\left[\diag(\lambda)-\eta\diag(d_z)^{-2}\right]W\diag(d_x)}_{M(\lambda,\eta)},l\Big) \implies M\succeq 0
\]
where $\gamma$ is the scaled Gershgorin operation defined in (\ref{eq:gershgorin_operation}). Since setting $\sigma = \gamma(M(\lambda,\eta),l)$ ensures $M\succeq 0$, (\ref{eq:general_dual}) can be relaxed by setting $\sigma = \gamma(M(\lambda,\eta),l)$  and observe that
\begin{align*}
\frac{1}{2}(x^TMx-\|\sigma\|_q)&=\frac{1}{2}\left(\diag(d_x)^{-1}x\right)^T\left(\diag(\sigma) - M(\lambda,\eta)\right)\left(\diag(d_x)^{-1}x\right)-\frac{1}{2}\|\sigma\|_q\\
&\geq -\frac{1}{2} \|\gamma(M(\lambda,\eta),l)\|_q.
\end{align*}
Hence, the diagonal-dominant surrogate yields the following relaxation for \eqref{eq:general_dual}
\begin{equation*}
 	\begin{aligned}
 		\max_{\substack{\mu,s,t,\tau,\lambda\in\R^m_+ \\l\in\R^n_{++},\eta\in\R_+}}\ & g^Tz-2\tau^T\qub-\frac{1}{2}\left(\eta+\|\gamma(M(\lambda,\eta),l)\|_q-\hat z^T\diag(\lambda)\hat z+\mu^T\1\right)\\
	\text{s.t. }\quad\ & g = \frac{1}{2}c + s - t - (2\pub-1)\odot\tau - \lambda\odot\hat z, \quad \phi = \frac{1}{2}c-s-t+\tau,\\
  		&\mu_j\lambda_j \geq  \phi_j^2\quad\text{for $j=1,\ldots,m$}.
 	\end{aligned}
\end{equation*}
The desired result in Corollary~\ref{cor:sdp_crown} follows by applying (\ref{eq:st}) to eliminate $s$ and $t$, and then optimize over $\mu$.

\newpage
\section{Derivation for the dual problem}\label{app:dual}
We describe how to derive the dual problem \eqref{eq:ellipsoid_dual}.  Recall that we have the primal problem
\[	
\min_{\substack{x\in\R^n,X\in\S^n\\
z,r,R\in\R^m}}\frac{1}{2}c^T(z+r)\quad \text{s.t.}\quad
\begin{array}{c}
	|z|\leq r \leq (2\pub-1)\odot z + 2\qub, \quad z = Wx+\hat z,\\
R \leq \diag(WXW^T) + 2z\odot \hat z - \hat z\odot \hat z,\\
\tr(\diag(d_x)^{-2}X)\leq 1,\quad \tr(\diag(d_z)^{-2}WXW^T)\leq 1,\\
 \begin{bmatrix}
	1 & x^T \\ x & X
\end{bmatrix}\succeq 0,\quad \begin{bmatrix}
	1 & r_j \\ r_j & R_j
\end{bmatrix}\succeq 0\quad\text{for $j=1,\ldots,m$}
\end{array}
\]

Let $s,t,\tau\in\R^m_+$ denote the dual variables for $z\leq r$, $-z\leq r$ and $r \leq (2\pub-1)\odot z + 2\qub$, respectively. Let $\lambda\in\R^m_+$ denote the dual variable of $R \leq \diag(WXW^T) + 2z\odot \hat z - \hat z\odot \hat z$. Let $\sigma,\eta\geq 0$ denote the dual variables for $\tr(\diag(d_x)^{-2}X) \leq 1$ and $\tr(\diag(d_z)^{-2}WXW^T)\leq 1$, respectively. Finally, let $\begin{bmatrix}\nu&y^T\\y&\tilde X\end{bmatrix}\succeq 0$ and $\begin{bmatrix}\mu_i&\tilde r_i^T\\\tilde r_i&\tilde R_i\end{bmatrix}\succeq 0$ denote the dual variables for the PSD blocks $\begin{bmatrix}
	1 & x^T \\ x & X
\end{bmatrix}\succeq 0,\  \begin{bmatrix}
	1 & r_i \\ r_i & R_i
\end{bmatrix}\succeq 0$, respectively. The Lagrangian is given by:
\begin{align*}
&-2\tau^T\qub-\sigma-\eta+\hat z^T\diag(\lambda)\hat z-\mu^T\1-\nu\\
+&\inner{Wx+\hat z}{\underbrace{\frac{1}{2}c+s-t-\tau\odot(2\pub-1)-2\lambda\odot\hat z}_{g}}-\inner{2y}{x}\\
+&\inner{X}{\sigma \diag(d_x)^{-2}- W^T\left[\diag(\lambda)-\eta\diag(d_z)^{-2}\right]W-\tilde X}\\
+&\inner{r}{\underbrace{\frac{1}{2}c-s-t+\tau}_{\phi}}-\inner{r}{2\tilde r}
+\inner{R}{\lambda-\tilde R}.
\end{align*}
Fixing $x$ and $z=Wx+\hat z$, and taking Lagrangian dual yield
\[
\phi-2\tilde r=0,\quad \lambda-\tilde R=0,\quad \sigma\diag(d_x)^{-2}- W^T\left[\diag(\lambda)-\eta\diag(d_z)^{-2}\right]W-\tilde X=0,
\]
which gives
\begin{equation*}
 	\begin{aligned}
 		\max_{\substack{\mu,s,t,\tau,\lambda\in\R^m_+ \\y\in\R^n, \sigma,\eta,\nu\in\R_+}}\ & g^Tz-2y^Tx-2\tau^T\qub-\sigma-\eta+\hat z^T\diag(\lambda)\hat z-\mu^T\1-\nu\\
 		\text{s.t. }\ \ \ & g = \frac{1}{2}c + s - t - (2\pub-1)\odot\tau - 2\lambda\odot\hat z, \quad\phi = \frac{1}{2}c-s-t+\tau,\\
 		& M = \sigma\diag(d_x)^{-2}- W^T\left[\diag(\lambda)-\eta\diag(d_z)^{-2}\right]W,\\
 		&\begin{bmatrix}\nu& y^T\\y&M\end{bmatrix}\succeq 0,\quad
 		\begin{bmatrix}\mu_j& \phi_j/2\\\phi_j/2&\lambda_j\end{bmatrix}\succeq 0\quad\text{for $j=1,\ldots,m$}.
 	\end{aligned}
\end{equation*}
It then follows from the identity that for any $M\succeq 0$
\[
x^TMx=\max_{\nu,y}\ -2y^Tx - \nu\quad\text{s.t.}\quad \begin{bmatrix}
	\nu&y^T\\y&M
\end{bmatrix}\succeq 0
\]
that $\nu$ and $y$ can be eliminated to yield
\begin{equation*}
 	\begin{aligned}
 		\max_{\substack{\mu,s,t,\tau,\lambda\in\R^m_+ \\\sigma,\eta\in\R_+}}\ & g^Tz+x^TMx-2\tau^T\qub-\sigma-\eta+\hat z^T\diag(\lambda)\hat z-\mu^T\1\\
 		\text{s.t. }\ \ \ & g = \frac{1}{2}c + s - t - (2\pub-1)\odot\tau - 2\lambda\odot\hat z, \quad\phi = \frac{1}{2}c-s-t+\tau,\\
 		& M = \sigma\diag(d_x)^{-2}- W^T\left[\diag(\lambda)-\eta\diag(d_z)^{-2}\right]W\succeq 0,\\
 		&\begin{bmatrix}\mu_j& \phi_j/2\\\phi_j/2&\lambda_j\end{bmatrix}\succeq 0\quad\text{for $j=1,\ldots,m$}.
 	\end{aligned}
\end{equation*}
Rescaling $\sigma\equiv \frac{1}{2}\sigma$, $\eta\equiv \frac{1}{2}\eta$, $\lambda\equiv \frac{1}{2}\lambda$ and $\mu\equiv\frac{1}{2}\mu$ to yield
\begin{equation*}
 	\begin{aligned}
 		\max_{\substack{\mu,s,t,\tau,\lambda\in\R^m_+ \\\sigma,\eta\in\R_+}}\ & g^Tz-2\tau^T\qub+\frac{1}{2}x^TMx-\frac{1}{2}\left(\sigma+\eta-\hat z^T\diag(\lambda)\hat z+\mu^T\1\right)\\
 		\text{s.t. }\ \ \ & g = \frac{1}{2}c + s - t - (2\pub-1)\odot\tau - \lambda\odot\hat z, \quad \phi = \frac{1}{2}c-s-t+\tau,\\
 		& M = \sigma\diag(d_x)^{-2}- W^T\left[\diag(\lambda)-\eta\diag(d_z)^{-2}\right]W\succeq 0,\\
 		&\begin{bmatrix}\mu_j& \phi_j\\\phi_j&\lambda_j\end{bmatrix}\succeq 0\quad\text{for $j=1,\ldots,m$}.
 	\end{aligned}
\end{equation*}
Notice that for a symmetric $2\times 2$ matrix, $X\succeq 0$ if and only if $\det(X)\geq 0$ and $\tr(X)\geq 0$. Applying this insight yields the desired dual in \eqref{eq:ellipsoid_dual}
\begin{equation*}
 \begin{aligned}
	\max_{\substack{\mu,s,t,\tau,\lambda\in\R^m_+ \\\sigma,\eta\in\R_+}}\ & g^Tz-2\tau^T\qub+\frac{1}{2}x^TMx-\frac{1}{2}\left(\sigma+\eta-\hat z^T\diag(\lambda)\hat z+\mu^T\1\right)\\
	\text{s.t. }\quad\ & g = \frac{1}{2}c + s - t - (2\pub-1)\odot\tau - \lambda\odot\hat z, \quad \phi = \frac{1}{2}c-s-t+\tau,\\
	& M = \sigma\diag(d_x)^{-2}- W^T\left[\diag(\lambda)-\eta\diag(d_z)^{-2}\right]W\succeq 0,\\
	&\mu_j\lambda_j\geq \phi_j^2\quad\text{for $j=1,\ldots,m$}.
 \end{aligned}
\end{equation*}

\section{Experimental details}\label{app:model_details}

\subsection{Computational resources}
All experiments were run on a server equipped with one AMD engineering-sample CPU (32 physical cores, 64 threads, 128 MB L3 cache) and four NVIDIA RTX 4090 GPUs.

\subsection{Model architecture}
The Lipschitz-regularized networks are taken directly from the previous SDP-CROWN~\citep{chiu2025sdp}, whereas the non-Lipschitz-regularized networks are taken from~\citep{salman2019convex,Dathathri2020SDP}. Detailed model architectures are listed in  Table~\ref{tab:model_architectures}. We also provided naive Lipschitz constant estimation by multiplying the spectral norm of each weight matrix. Several convolutional layers in Table~\ref{tab:model_architectures} exhibit a diagonal-dominant structure.

\begin{figure}[b]
	\center
	\includegraphics[width=0.4\textwidth]{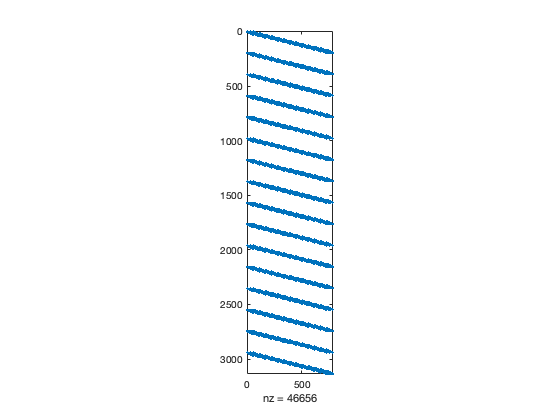}%
	\includegraphics[width=0.4\textwidth]{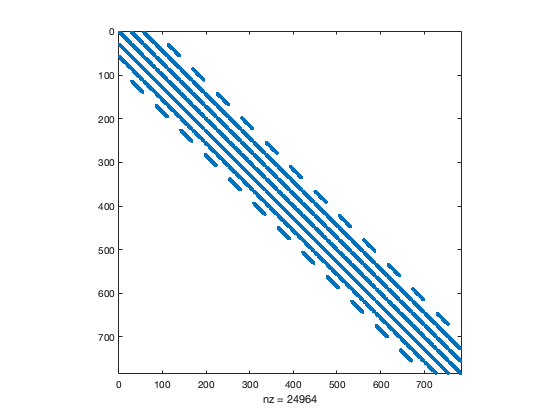}
	\caption{\textbf{Diagonal-dominant structure in convolutional layers.} \textbf{(Left.)} The sparsity pattern of the weight matrix $W$ for Conv(1, 16, 4, 2, 1), which is the first layer of CNN-small NOR (MNIST) and CNN-small ADV (MNIST) in Table~\ref{tab:model_architectures}. \textbf{(Right.)} The sparsity pattern of $W^TW$ exhibits a diagonal-dominant structure, with nonzero entries concentrated near the diagonal. The matrix $M(\lambda,\eta)$ inherits this banded sparsity structure.}\label{fig:dd}
\end{figure}

\begin{table*}[t]
\centering
\caption{\textbf{Models used in our experiments.} Conv($c_{\mathrm{in}}$, $c_{\mathrm{out}}$, $k$, $s$, $p$) denotes a convolutional layer with $c_{\mathrm{in}}$ input channels, $c_{\mathrm{out}}$ output channels, kernel size $k \times k$, stride $s$, and padding $p$. Linear($d_{\mathrm{in}}$, $d_{\mathrm{out}}$) denotes a fully connected layer with $d_{\mathrm{in}}$ input features and $d_{\mathrm{out}}$ output features. ZeroPad2d$(p_l, p_r, p_t, p_b)$ adds $p_l$, $p_r$, $p_t$, and $p_b$ pixels of zero padding to the left, right, top, and bottom of each input channel, respectively. There are no max-pooling or average-pooling layers in these models, and ReLU activations are applied between consecutive layers, except after ZeroPad2d.}
\label{tab:model_architectures}
\renewcommand{\arraystretch}{1.18}
\resizebox{\textwidth}{!}{
\begin{tabular}{l|p{0.56\textwidth}|c|c}
\toprule
\textbf{Model name} & \textbf{Model structure} & \textbf{Accuracy} & \textbf{Lipschitz} \\
\midrule

\multicolumn{4}{c}{\textbf{Small Lipschitz-constant networks}} \\
\midrule
\multicolumn{4}{c}{\textbf{MNIST Models}} \\
\midrule

MLP (MNIST) 
& Linear(784, 100) - Linear(100, 100) - Linear(100, 10) 
& 79\% 
& 0.98 \\
\midrule

ConvSmall (MNIST) 
& Conv(1, 16, 4, 2, 1) - Conv(16, 32, 4, 2, 1) - Linear(1568, 100) - Linear(100, 10) 
& 87.5\% 
& 1.14 \\
\midrule

ConvLarge (MNIST) 
& Conv(1, 32, 3, 1, 1) - Conv(32, 32, 4, 2, 1) - Conv(32, 64, 3, 1, 1) - Conv(64, 64, 4, 2, 1) - Linear(3136, 512) - Linear(512, 512) - Linear(512, 10) 
& 89.5\% 
& 1.21 \\
\midrule

\multicolumn{4}{c}{\textbf{CIFAR-10 Models}} \\
\midrule

CNN-A (CIFAR-10) 
& Conv(3, 16, 4, 2, 1) - Conv(16, 32, 4, 2, 1) - Linear(2048, 100) - Linear(100, 10) 
& 62\% 
& 0.99 \\
\midrule

CNN-B (CIFAR-10) 
& ZeroPad2d(1, 2, 1, 2) - Conv(3, 32, 5, 2, 0) - Conv(32, 128, 4, 2, 1) - Linear(8192, 250) - Linear(250, 10) 
& 66\% 
& 1.53 \\
\midrule

CNN-C (CIFAR-10) 
& Conv(3, 8, 4, 2, 0) - Conv(8, 16, 4, 2, 0) - Linear(576, 128) - Linear(128, 64) - Linear(64, 10) 
& 51\% 
& 1.06 \\
\midrule

ConvSmall (CIFAR-10) 
& Conv(3, 16, 4, 2, 0) - Conv(16, 32, 4, 2, 0) - Linear(1152, 100) - Linear(100, 10) 
& 60.5\% 
& 1.52 \\
\midrule

ConvDeep (CIFAR-10) 
& Conv(3, 8, 4, 2, 1) - Conv(8, 8, 3, 1, 1) - Conv(8, 8, 3, 1, 1) - Conv(8, 8, 4, 2, 1) - Linear(512, 100) - Linear(100, 10) 
& 53\% 
& 0.79 \\
\midrule

ConvLarge (CIFAR-10) 
& Conv(3, 32, 3, 1, 1) - Conv(32, 32, 4, 2, 1) - Conv(32, 64, 3, 1, 1) - Conv(64, 64, 4, 2, 1) - Linear(4096, 512) - Linear(512, 512) - Linear(512, 10) 
& 74\% 
& 5.13 \\
\midrule

\multicolumn{4}{c}{\textbf{Large Lipschitz-constant networks}} \\
\midrule
\multicolumn{4}{c}{\textbf{MNIST Models}} \\
\midrule

CNN-small NOR (MNIST) 
& Conv(1, 16, 4, 2, 1) - Conv(16, 32, 4, 2, 1) - Linear(1568, 100) - Linear(100, 10) 
& 98.5\% 
& 1132.08 \\
\midrule

CNN-small ADV (MNIST) 
& Conv(1, 16, 4, 2, 1) - Conv(16, 32, 4, 2, 1) - Linear(1568, 100) - Linear(100, 10) 
& 98.5\% 
& 513.95 \\
\midrule

\multicolumn{4}{c}{\textbf{CIFAR-10 Models}} \\
\midrule

CNN-small NOR (CIFAR-10) 
& Conv(3, 16, 4, 2, 1) - Conv(16, 32, 4, 2, 1) - Linear(2048, 100) - Linear(100, 10)
& 80.5\% 
& 325.31 \\
\midrule

CNN-A-Adv (CIFAR-10) 
& Conv(3, 16, 4, 2, 1) - Conv(16, 32, 4, 2, 1) - Linear(2048, 100) - Linear(100, 10) 
& 70.5\%
& 632.41 \\
\midrule

CNN-A-Adv4 (CIFAR-10) 
& Conv(3, 16, 4, 2, 1) - Conv(16, 32, 4, 2, 1) - Linear(2048, 100) - Linear(100, 10) 
& 60.5\% 
& 445.97 \\

\bottomrule
\end{tabular}
}
\end{table*}

\section{Limitations}\label{app:limitation}
Our method currently supports only ReLU networks. Extending it to general activations and architectures remains future work. Although Theorem~\ref{thm:sdp_crown} permits general affine-image enclosures, our implementation uses each layer's weight matrix as $W$ and inherits approximation error from the preceding layer. Efficiently constructing and certifying tighter enclosures using information from multiple layers remains a future direction.


\newpage
\input{checklist.tex}

\end{document}

%% file: sections/bab.tex
Our method can also be integrated into the branch-and-bound framework of $\beta$-CROWN~\citep{wang2021beta}. We retain the $\alpha,\beta$-CROWN branch-and-bound framework and replace only the bound computation for each subproblem with our approach. Each ReLU split introduces an additional dual variable $\beta$, which is then optimized jointly with the other multipliers of our bound. A subproblem inherits and is warm-started with the optimized multipliers of its parent, so a few optimization steps suffice per subproblem. As a design choice, we fix the affine-image multipliers $\lambda$, including the scaling vector $l$, at their optimized root values throughout branch-and-bound. This choice allows the affine-image term, the most expensive component of the bound, to be computed once rather than at every optimization step.

%% file: sections/experiment.tex
\paragraph{Benchmarks.} We evaluate our method on both Lipschitz-regularized networks and non-Lipschitz-regularized networks. The Lipschitz-regularized networks are adopted directly from the previous SDP-CROWN paper~\citep{chiu2025sdp}, using the same perturbation radii and experimental settings. To further test scalability and robustness on more challenging models, we also evaluate on non-Lipschitz-regularized networks from prior works~\citep{salman2019convex,Dathathri2020SDP}. Details of the model architectures and estimated Lipschitz constants are provided in Appendix~\ref{app:model_details}.

\paragraph{Baselines \& Implementations.}  We compare our method with several state-of-the-art
verifiers for robustness under $\ell_2$-norm perturbations. In particular, we consider two incomplete verifiers, namely $\alpha$-CROWN~\citep{xu2021fast} and the original SDP-CROWN~\citep{chiu2025sdp}, which restricts the linear coefficient $g$ to the $\alpha$-CROWN parametrization. We also consider two stronger verifiers that incorporate branch-and-bound and cutting planes, $\beta$-CROWN~\citep{wang2021beta} and BICCOS~\citep{zhou2024scalable}. We intentionally omit GCP-CROWN in our comparison as its cut generation process cannot generate proper cuts for $\ell_2$ perturbations. We also include LipNaive, a simple Lipschitz
baseline used in previous work, which certifies robustness with an upper bound on
the network's Lipschitz constant obtained by multiplying the Lipschitz constants
of the individual weight matrices. On the Lipschitz-regularized networks, both the
original SDP-CROWN and our affine-image SDP-CROWN use the ball propagated with
the Lipschitz constants, as in the original SDP-CROWN paper, since this ball is
already nearly tight on these networks. On the non-Lipschitz-regularized networks, our
affine-image SDP-CROWN additionally imposes the affine-image constraint on the
first layer, with the scaling $l$ optimized jointly,
and the original SDP-CROWN is run with the elementwise interval constraint extension described in Theorem~C.1 of the original paper~\citep{chiu2025sdp}. We report our method both on its own and within branch-and-bound
(Affine-Image SDP-CROWN + BaB). For all branch-and-bound-based verification methods, we use a 150-second timeout per instance.

\begin{table*}[t]
\centering
\small
\setlength{\tabcolsep}{3.5pt}
\renewcommand{\arraystretch}{1.5}
\resizebox{\textwidth}{!}{
\begin{tabular}{lcccccccc}
\toprule
Network &
\begin{tabular}[c]{@{}c@{}}Upper\\
Bound\end{tabular} &
\begin{tabular}[c]{@{}c@{}}Affine-Image\\
SDP-CROWN + BaB\end{tabular} &
\begin{tabular}[c]{@{}c@{}}Affine-Image\\
SDP-CROWN\end{tabular} &
\begin{tabular}[c]{@{}c@{}}Original\\
SDP-CROWN\end{tabular} &
BICCOS &
$\beta$-CROWN &
$\alpha$-CROWN &
LipNaive \\
\midrule
\multicolumn{9}{c}{\textbf{Lipschitz-regularized /
Small Lipschitz-constant Networks}} \\
\midrule
\multicolumn{9}{c}{\textbf{MNIST Models}} \\
\midrule
MLP $(\rho=1.0)$
  & 50.5\%
  & \textbf{49.5\%} (8.36s)
  & 44\% (2.10s)
  & 32.5\% (1.93s)
  & 40.5\% (57.51s)
  & 39\% (48.87s)
  & 1.5\% (1.38s)
  & 29\% (0.02s) \\
ConvSmall $(\rho=0.3)$
  & 84.5\%
  & \textbf{84.5\%} (1.91s)
  & 84\% (1.81s)
  & 81.5\% (5.15s)
  & 17\% (128.72s)
  & 16.5\% (124.74s)
  & 0\% (2.39s)
  & 77.5\% (0.10s) \\
ConvLarge $(\rho=0.3)$
  & 84\%
  & \textbf{82.5\%} (7.36s)
  & \textbf{82.5\%} (5.05s)
  & 79.5\% (32.71s)
  & 0\% (150.00s)
  & 0\% (150.00s)
  & 0\% (8.21s)
  & 77\% (1.00s) \\
\midrule
\multicolumn{9}{c}{\textbf{CIFAR-10 Models}} \\
\midrule
CNN-A $(\rho=\frac{24}{255})$
  & 55.5\%
  & \textbf{54\%} (8.70s)
  & 51.5\% (2.85s)
  & 49\% (4.75s)
  & 20\% (97.59s)
  & 20\% (97.25s)
  & 7.5\% (2.20s)
  & 39\% (0.20s) \\
CNN-B $(\rho=\frac{24}{255})$
  & 59.5\%
  & \textbf{58\%} (10.99s)
  & 52.5\% (6.65s)
  & 49.5\% (9.02s)
  & 3\% (143.30s)
  & 3\% (142.71s)
  & 0\% (3.20s)
  & 33\% (0.30s) \\
CNN-C $(\rho=\frac{24}{255})$
  & 47\%
  & \textbf{45.5\%} (7.10s)
  & 45\% (3.30s)
  & 42.5\% (4.76s)
  & 35.5\% (39.98s)
  & 35.5\% (38.26s)
  & 24.5\% (1.91s)
  & 36.5\% (0.20s) \\
ConvSmall $(\rho=\frac{24}{255})$
  & 52.5\%
  & \textbf{51\%} (8.04s)
  & 47.5\% (3.58s)
  & 43.5\% (3.94s)
  & 18\% (102.03s)
  & 18\% (100.94s)
  & 6\% (2.30s)
  & 33\% (0.20s) \\
ConvDeep $(\rho=\frac{24}{255})$
  & 50\%
  & \textbf{49.5\%} (7.06s)
  & 47.5\% (5.73s)
  & 45.5\% (11.19s)
  & 31.5\% (59.39s)
  & 30.5\% (59.87s)
  & 22.5\% (2.53s)
  & 39.5\% (0.30s) \\
ConvLarge $(\rho=\frac{8}{255})$
  & 72.5\%
  & \textbf{70.5\%} (14.27s)
  & 66.5\% (7.72s)
  & 63.5\% (38.51s)
  & 6\% (139.77s)
  & 6\% (139.02s)
  & 2.5\% (11.08s)
  & 47.5\% (1.20s) \\
\midrule
\multicolumn{9}{c}{\textbf{Non-Lipschitz regularized /
Large Lipschitz-constant networks}} \\
\midrule
\multicolumn{9}{c}{\textbf{MNIST Models}} \\
\midrule
CNN-small NOR $(\rho = 0.15)$
  & 98\%
  & \textbf{86.5\%} (26.67s)
  & 69\% (10.07s)
  & 42\% (8.39s)
  & 30.5\% (111.96s)
  & 29.5\% (108.78s)
  & 3\% (2.34s)
  & 0\% (0.40s) \\
CNN-small ADV $(\rho = 0.25)$
  & 98.5\%
  & \textbf{96.5\%} (6.99s)
  & 90.5\% (4.39s)
  & 33\% (7.80s)
  & 3.5\% (146.96s)
  & 2.5\% (147.47s)
  & 0\% (2.46s)
  & 0\% (0.40s) \\
\midrule
\multicolumn{9}{c}{\textbf{CIFAR-10 Models}} \\
\midrule
CNN-small NOR $(\rho = \frac{8}{255})$
  & 75.5\%
  & \textbf{61.5\%} (40.28s)
  & 51.5\% (21.05s)
  & 34.5\% (7.42s)
  & 25\% (103.88s)
  & 24\% (104.56s)
  & 5\% (2.33s)
  & 0\% (0.36s) \\
CNN-A-Adv $(\rho = \frac{32}{255})$
  & 63.5\%
  & \textbf{49\%} (49.05s)
  & 39.5\% (24.57s)
  & 9\% (8.04s)
  & 13.5\% (122.75s)
  & 12.5\% (122.15s)
  & 0.5\% (2.51s)
  & 0\% (0.40s) \\
CNN-A-Adv4 $(\rho = \frac{32}{255})$
  & 53\%
  & \textbf{42\%} (42.96s)
  & 37.5\% (20.90s)
  & 14.5\% (7.32s)
  & 13.5\% (116.38s)
  & 12.5\% (117.83s)
  & 0.5\% (2.54s)
  & 0\% (0.36s) \\
\bottomrule
\end{tabular}
}
\caption{\textbf{Verified accuracy and average runtime
on benchmarks.}
Bold entries indicate the best verified accuracy.
The perturbation size is indicated next to the network name.
We can see that our proposed method shows superior
performance in all the benchmarks. Impressively, our affine-image SDP-CROWN without branch-and-bound already surpasses strong verifiers that incorporate branch-and-bound on these benchmarks.}
\label{tab:main_table}
\end{table*}

\subsection{Verification accuracy comparison}

Table~\ref{tab:main_table} reports the verified accuracy and average runtime of our method and the baseline verifiers under $\ell_2$ norm perturbation. The ``Upper Bound'' column is computed using PGD attacks and upper bounds the maximum achievable verified accuracy. We first evaluate on the Lipschitz-regularized networks used in the previous SDP-CROWN paper. On these benchmarks, our incomplete verifier consistently improves over SDP-CROWN at comparable computational cost, increasing verified accuracy by $2$--$11.5\%$ across all benchmarks. When integrated with branch-and-bound, our method achieves saturating performance on most benchmarks. Compared with other branch-and-bound-based verifiers such as BICCOS and $\beta$-CROWN, our method is often substantially faster while achieving higher certified accuracy, demonstrating a favorable accuracy-efficiency tradeoff.

We further evaluate on non-Lipschitz-regularized networks, which are more challenging for the Lipschitz-based method and also the original SDP-CROWN. On these models, the gap between our method and previous methods becomes more pronounced. Although the original SDP-CROWN can still significantly improve over $\alpha$-CROWN, the performance is still not as competitive as in the Lipschitz-regularized case, and LipNaive fails completely. In contrast, our method certifies nontrivial robustness across all tested models, achieving close-to-saturating verified accuracy on the MNIST CNN-small NOR and ADV models. On CIFAR-10, our method with branch and bound certifies $61.5\%$ on CNN-small NOR and improves over $\beta$-CROWN and BICCOS by more than $3$ times on CNN-A-Adv4 and CNN-A-Adv. These results suggest that our method is substantially more robust than previous bound-propagation methods on non-Lipschitz-regularized networks when considering $l_2$ perturbations, where the original SDP-CROWN relaxation becomes overly conservative.

\subsection{Lower bound comparison}

\begin{figure}[t]
    \centering
    \includegraphics[width=\linewidth]{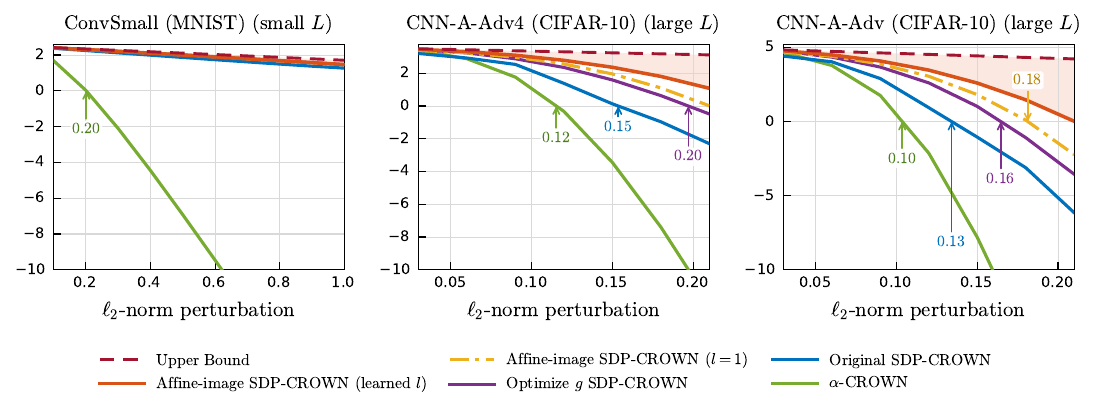}
    \caption{\textbf{Lower bounds on the classification margin against $\ell_2$ perturbations.} We compare lower bounds on $c^T f(x)$ obtained using our full affine-image SDP-CROWN, affine-image SDP-CROWN with fixed scaling $l=1$, the original SDP-CROWN with optimized $g$, the original SDP-CROWN that optimizes only the bias term, and $\alpha$-CROWN, together with a PGD-based empirical upper bound. The lower bounds are averaged over 20 instances and all 9 incorrect specifications. The annotated numbers indicate where each lower-bound curve intersects zero, corresponding to the average perturbation radius certified by each method. Small $L$ corresponds to Lipschitz-regularized networks, whereas large $L$ corresponds to networks without Lipschitz regularization.}
    \label{fig:l2_margin_curves}
\end{figure}

We further analyze the lower bounds produced by our method and baselines over a wide range of $\ell_2$ perturbation radii, as reported in Figure~\ref{fig:l2_margin_curves}. We also include an ablation that compares our full configuration with two reduced variants, one that optimizes the slope $g$ without the affine image, and one that uses the affine image with the scaling fixed at $l = 1$. For each perturbation radius, we compute certified lower bounds on the classification margins $c^T f(x)$. To reduce randomness, we take the average lower bound over 20 input instances and all 9 incorrect-class specifications for each instance. For comparison, we also added an upper bound obtained from projected gradient descent (PGD). As shown in Figure~\ref{fig:l2_margin_curves}, our method has a small advantage over the original SDP-CROWN in Lipschitz-regularized networks. And since the propagated ball is already nearly tight, the ablation variants of our methods essentially coincide. On non-Lipschitz-regularized networks, the advantage of our setting is more pronounced. Although much tighter than $\alpha$-CROWN, the bounds produced by the original SDP-CROWN is still conservative. In contrast, our proposed method can consistently produce much tighter bounds than both the original SDP-CROWN and $\alpha$-CROWN, as reflected by both the higher margin curves and the larger zero-crossing radii. In addition, these results reveal our hierarchy more clearly. Optimizing $g$ in the original SDP-CROWN reduces the functional loss and tightens the bounds. The affine-image formulation further reduces the geometric loss, yielding even tighter bounds. Finally, jointly optimizing the scaling factor $l$ within the affine-image formulation further reduces the geometric loss and yields the tightest bounds. Finally, the improvement over the baselines grows as the perturbation radius increases. While all methods are relatively close at small radii, the bounds from $\alpha$-CROWN and original SDP-CROWN degrade much faster than those from our method. Overall, these results demonstrate that our method provides a significantly tighter relaxation and yields stronger certified robustness margins under $\ell_2$ perturbations.

%% file: checklist.tex
\section*{NeurIPS Paper Checklist}

\begin{enumerate}

\item {\bf Claims}
    \item[] Question: Do the main claims made in the abstract and introduction accurately reflect the paper's contributions and scope?
    \item[] Answer: \answerYes{} 
    \item[] Justification: We clearly stated our contirbution in the abstract and introduction. The contribution of this paper is to develop novel bound propagation method for neural network verification.

\item {\bf Limitations}
    \item[] Question: Does the paper discuss the limitations of the work performed by the authors?
    \item[] Answer: \answerYes{}. 
    \item[] Justification: Limitations are discussed in Appendix~\ref{app:limitation}.

\item {\bf Theory assumptions and proofs}
    \item[] Question: For each theoretical result, does the paper provide the full set of assumptions and a complete (and correct) proof?
    \item[] Answer: \answerYes{} 
    \item[] Justification: We provide a complete proof for all the Theorem in this paper.

    \item {\bf Experimental result reproducibility}
    \item[] Question: Does the paper fully disclose all the information needed to reproduce the main experimental results of the paper to the extent that it affects the main claims and/or conclusions of the paper (regardless of whether the code and data are provided or not)?
    \item[] Answer: \answerYes{} 
    \item[] Justification: We clearly state all the experimental setting in the paepr. 

\item {\bf Open access to data and code}
    \item[] Question: Does the paper provide open access to the data and code, with sufficient instructions to faithfully reproduce the main experimental results, as described in supplemental material?
    \item[] Answer: \answerYes{} 
    \item[] Justification: We plan to release the full implementation and experimental code upon acceptance of the paper. We are also willing to provide additional implementation details, experimental settings, and clarification regarding the evaluation protocol should the reviewers request them.

\item {\bf Experimental setting/details}
    \item[] Question: Does the paper specify all the training and test details (e.g., data splits, hyperparameters, how they were chosen, type of optimizer) necessary to understand the results?
    \item[] Answer: \answerYes{} 
    \item[] Justification: All the experimental setting are specified.

\item {\bf Experiment statistical significance}
    \item[] Question: Does the paper report error bars suitably and correctly defined or other appropriate information about the statistical significance of the experiments?
    \item[] Answer: \answerNo{} 
    \item[] Justification: There are no error bars to be reported for our main experiments.
    \item[] Guidelines: All the information about the statistical significance of our experiments are correct defined.
    \begin{itemize}
        \item The answer \answerNA{} means that the paper does not include experiments.
        \item The authors should answer \answerYes{} if the results are accompanied by error bars, confidence intervals, or statistical significance tests, at least for the experiments that support the main claims of the paper.
        \item The factors of variability that the error bars are capturing should be clearly stated (for example, train/test split, initialization, random drawing of some parameter, or overall run with given experimental conditions).
        \item The method for calculating the error bars should be explained (closed form formula, call to a library function, bootstrap, etc.)
        \item The assumptions made should be given (e.g., Normally distributed errors).
        \item It should be clear whether the error bar is the standard deviation or the standard error of the mean.
        \item It is OK to report 1-sigma error bars, but one should state it. The authors should preferably report a 2-sigma error bar than state that they have a 96\% CI, if the hypothesis of Normality of errors is not verified.
        \item For asymmetric distributions, the authors should be careful not to show in tables or figures symmetric error bars that would yield results that are out of range (e.g., negative error rates).
        \item If error bars are reported in tables or plots, the authors should explain in the text how they were calculated and reference the corresponding figures or tables in the text.
    \end{itemize}

\item {\bf Experiments compute resources}
    \item[] Question: For each experiment, does the paper provide sufficient information on the computer resources (type of compute workers, memory, time of execution) needed to reproduce the experiments?
    \item[] Answer: \answerYes{} 
    \item[] Justification: We provide detailed information on the CPU/GPUs being used in our study in the Appendix~\ref{app:model_details}.
    
\item {\bf Code of ethics}
    \item[] Question: Does the research conducted in the paper conform, in every respect, with the NeurIPS Code of Ethics \url{https://neurips.cc/public/EthicsGuidelines}?
    \item[] Answer: \answerYes{} 
    \item[] Justification: The research conducted in the paper conform, in every respect, with the NeurIPS Code of Ethics

\item {\bf Broader impacts}
    \item[] Question: Does the paper discuss both potential positive societal impacts and negative societal impacts of the work performed?
    \item[] Answer: \answerNA{} 
    \item[] Justification: There is no societal impact of the work performed. The study is very theoretical, and we do not foresee any potential harm to society.
    
\item {\bf Safeguards}
    \item[] Question: Does the paper describe safeguards that have been put in place for responsible release of data or models that have a high risk for misuse (e.g., pre-trained language models, image generators, or scraped datasets)?
    \item[] Answer: \answerNA{} 
    \item[] Justification: This paper poses no such risks.

\item {\bf Licenses for existing assets}
    \item[] Question: Are the creators or original owners of assets (e.g., code, data, models), used in the paper, properly credited and are the license and terms of use explicitly mentioned and properly respected?
    \item[] Answer: \answerYes{} 
    \item[] Justification: All the assets used in the paper are properly cited.

\item {\bf New assets}
    \item[] Question: Are new assets introduced in the paper well documented and is the documentation provided alongside the assets?
    \item[] Answer: \answerNA{} 
    \item[] Justification: This paper does not release new assets.

\item {\bf Crowdsourcing and research with human subjects}
    \item[] Question: For crowdsourcing experiments and research with human subjects, does the paper include the full text of instructions given to participants and screenshots, if applicable, as well as details about compensation (if any)? 
    \item[] Answer: \answerNA{} 
    \item[] Justification: This paper does not involve crowdsourcing nor research with human subjects.

\item {\bf Institutional review board (IRB) approvals or equivalent for research with human subjects}
    \item[] Question: Does the paper describe potential risks incurred by study participants, whether such risks were disclosed to the subjects, and whether Institutional Review Board (IRB) approvals (or an equivalent approval/review based on the requirements of your country or institution) were obtained?
    \item[] Answer: \answerNA{} 
    \item[] Justification: This paper does not involve crowdsourcing nor research with human subjects.

\item {\bf Declaration of LLM usage}
    \item[] Question: Does the paper describe the usage of LLMs if it is an important, original, or non-standard component of the core methods in this research? Note that if the LLM is used only for writing, editing, or formatting purposes and does \emph{not} impact the core methodology, scientific rigor, or originality of the research, declaration is not required.
    \item[] Answer: \answerNA{} 
    \item[] Justification: The core method development in this research does not involve LLMs as any important, original, or non-standard components.

\end{enumerate}

%% file: robustness.bib
@inproceedings{de2021scaling,
  title={Scaling the convex barrier with active sets},
  author={De Palma, Alessandro and Behl, Harkirat S and Bunel, Rudy and Torr, Philip and Kumar, M Pawan},
  booktitle={Proceedings of the ICLR 2021 Conference},
  year={2021},
  organization={Open Review}
}

@inproceedings{ferrari2022complete,
  title={Complete Verification via Multi-Neuron Relaxation Guided Branch-and-Bound},
  author={Ferrari, Claudio and Mueller, Mark Niklas and Jovanovi{\'c}, Nikola and Vechev, Martin},
  booktitle={International Conference on Learning Representations},
  year={2022}
}

@inproceedings{hashemi2021osip,
  title={Osip: Tightened bound propagation for the verification of relu neural networks},
  author={Hashemi, Vahid and Kouvaros, Panagiotis and Lomuscio, Alessio},
  booktitle={International Conference on Software Engineering and Formal Methods},
  pages={463--480},
  year={2021},
  organization={Springer}
}

@inproceedings{barrett2022certifiably,
  title={Certifiably robust variational autoencoders},
  author={Barrett, Ben and Camuto, Alexander and Willetts, Matthew and Rainforth, Tom},
  booktitle={International Conference on Artificial Intelligence and Statistics},
  pages={3663--3683},
  year={2022},
  organization={PMLR}
}

@inproceedings{wong2021learning,
  title={Learning perturbation sets for robust machine learning},
  author={Wong, Eric and Kolter, J Zico},
  booktitle={International Conference on Learning Representations},
  year={2021}
}

@inproceedings{batten2021efficient,
  title={Efficient Neural Network Verification via Layer-based Semidefinite Relaxations and Linear Cuts.},
  author={Batten, Ben and Kouvaros, Panagiotis and Lomuscio, Alessio and Zheng, Yang},
  booktitle={IJCAI},
  pages={2184--2190},
  year={2021}
}

@inproceedings{xu2021fast,
  title={Fast and Complete: Enabling Complete Neural Network Verification with Rapid and Massively Parallel Incomplete Verifiers},
  author={Xu, Kaidi and Zhang, Huan and Wang, Shiqi and Wang, Yihan and Jana, Suman and Lin, Xue and Hsieh, Cho-Jui},
  booktitle={International Conference on Learning Representation (ICLR)},
  year={2021}
}

@article{wang2021beta,
  title={Beta-crown: Efficient bound propagation with per-neuron split constraints for neural network robustness verification},
  author={Wang, Shiqi and Zhang, Huan and Xu, Kaidi and Lin, Xue and Jana, Suman and Hsieh, Cho-Jui and Kolter, J Zico},
  journal={Advances in Neural Information Processing Systems},
  volume={34},
  pages={29909--29921},
  year={2021}
}

@article{zhang2018crown,
  title={Efficient neural network robustness certification with general activation functions},
  author={Zhang, Huan and Weng, Tsui-Wei and Chen, Pin-Yu and Hsieh, Cho-Jui and Daniel, Luca},
  journal={Advances in neural information processing systems},
  volume={31},
  year={2018}
}

@article{singh2019beyond,
  title={Beyond the single neuron convex barrier for neural network certification},
  author={Singh, Gagandeep and Ganvir, Rupanshu and P{\"u}schel, Markus and Vechev, Martin},
  journal={Advances in Neural Information Processing Systems},
  volume={32},
  year={2019}
}

@article{salman2019convex,
  title={A convex relaxation barrier to tight robustness verification of neural networks},
  author={Salman, Hadi and Yang, Greg and Zhang, Huan and Hsieh, Cho-Jui and Zhang, Pengchuan},
  journal={Advances in Neural Information Processing Systems},
  volume={32},
  year={2019}
}

@inproceedings{dathathri2020enabling,
  title={Enabling certification of verification-agnostic networks via memory-efficient semidefinite programming},
  author={Dathathri, Sumanth and Dvijotham, Krishnamurthy and Kurakin, Alexey and Raghunathan, Aditi and Uesato, Jonathan and Bunel, Rudy and Shankar, Shreya and Steinhardt, Jacob and Goodfellow, Ian and Liang, Percy and others},
  booktitle={Advances in Neural Information Processing Systems},
  year={2020}
}

@inproceedings{zhang2020tightness,
  title={On the tightness of semidefinite relaxations for certifying robustness to adversarial examples},
  author={Zhang, Richard Y},
  booktitle={Advances in Neural Information Processing Systems},
  year={2020}
}

@inproceedings{dvijotham2020efficient,
  title={Efficient neural network verification with exactness characterization},
  author={Dvijotham, Krishnamurthy Dj and Stanforth, Robert and Gowal, Sven and Qin, Chongli and De, Soham and Kohli, Pushmeet},
  booktitle={Uncertainty in Artificial Intelligence},
  pages={497--507},
  year={2020},
  organization={PMLR}
}

@inproceedings{singh2018fast,
  title={Fast and effective robustness certification},
  author={Singh, Gagandeep and Gehr, Timon and Mirman, Matthew and P{\"u}schel, Markus and Vechev, Martin},
  booktitle={Advances in Neural Information Processing Systems},
  pages={10802--10813},
  year={2018}
}

@inproceedings{dvijotham2018dual,
  title={A Dual Approach to Scalable Verification of Deep Networks.},
  author={Dvijotham, Krishnamurthy and Stanforth, Robert and Gowal, Sven and Mann, Timothy A and Kohli, Pushmeet},
  booktitle={UAI},
  volume={1},
  pages={2},
  year={2018}
}

@inproceedings{ehlers2017formal,
  title={Formal verification of piece-wise linear feed-forward neural networks},
  author={Ehlers, Ruediger},
  booktitle={International Symposium on Automated Technology for Verification and Analysis},
  pages={269--286},
  year={2017},
  organization={Springer}
}

@article{raghunathan2018semidefinite,
  title={Semidefinite relaxations for certifying robustness to adversarial examples},
  author={Raghunathan, Aditi and Steinhardt, Jacob and Liang, Percy S},
  journal={Advances in Neural Information Processing Systems},
  volume={31},
  year={2018}
}

@article{zhang2022general,
  title={General Cutting Planes for Bound-Propagation-Based Neural Network Verification},
  author={Zhang, Huan and Wang, Shiqi and Xu, Kaidi and Li, Linyi and Li, Bo and Jana, Suman and Hsieh, Cho-Jui and Kolter, J Zico},
  journal={Advances in Neural Information Processing Systems},
  year={2022}
}

@article{xu2020automatic,
  title={Automatic perturbation analysis for scalable certified robustness and beyond},
  author={Xu, Kaidi and Shi, Zhouxing and Zhang, Huan and Wang, Yihan and Chang, Kai-Wei and Huang, Minlie and Kailkhura, Bhavya and Lin, Xue and Hsieh, Cho-Jui},
  journal={Advances in Neural Information Processing Systems},
  volume={33},
  pages={1129--1141},
  year={2020}
}

@article{singh2019abstract,
  title={An abstract domain for certifying neural networks},
  author={Singh, Gagandeep and Gehr, Timon and P{\"u}schel, Markus and Vechev, Martin},
  journal={Proceedings of the ACM on Programming Languages},
  volume={3},
  number={POPL},
  pages={1--30},
  year={2019},
  publisher={ACM New York, NY, USA}
}

@inproceedings{shi2024genbab,
  title={Neural Network Verification with Branch-and-Bound for General Nonlinearities},
  author={Shi, Zhouxing and Jin, Qirui and Kolter, Zico and Jana, Suman and Hsieh, Cho-Jui and Zhang, Huan},
  booktitle={International Conference on Tools and Algorithms for the Construction and Analysis of Systems},
  year={2025}
}

@inproceedings{brown2022unified,
  title={A unified view of SDP-based neural network verification through completely positive programming},
  author={Brown, Robin A and Schmerling, Edward and Azizan, Navid and Pavone, Marco},
  booktitle={International conference on artificial intelligence and statistics},
  pages={9334--9355},
  year={2022},
  organization={PMLR}
}

@inproceedings{chiu2023tight,
  title={Tight certification of adversarially trained neural networks via nonconvex low-rank semidefinite relaxations},
  author={Chiu, Hong-Ming and Zhang, Richard Y},
  booktitle={International Conference on Machine Learning},
  pages={5631--5660},
  year={2023},
  organization={PMLR}
}

@article{ma2020strengthened,
  title={Strengthened SDP verification of neural network robustness via non-convex cuts},
  author={Ma, Ziye and Sojoudi, Somayeh},
  journal={arXiv preprint arXiv:2010.08603},
  pages={715--727},
  year={2020}
}

@article{fazlyab2020safety,
  title={Safety verification and robustness analysis of neural networks via quadratic constraints and semidefinite programming},
  author={Fazlyab, Mahyar and Morari, Manfred and Pappas, George J},
  journal={IEEE Transactions on Automatic Control},
  volume={67},
  number={1},
  pages={1--15},
  year={2020},
  publisher={IEEE}
}

@inproceedings{zhou2024scalable,
  title={Scalable Neural Network Verification with Branch-and-bound Inferred Cutting Planes},
  author={Zhou, Duo and Brix, Christopher and Hanasusanto, Grani A and Zhang, Huan},
  booktitle={The Thirty-eighth Annual Conference on Neural Information Processing Systems},
  year={2024}
}

@inproceedings{Dathathri2020SDP,
  author    = {S. Dathathri and K. Dvijotham and A. Kurakin and A. Raghunathan and J. Uesato and R. R. Bunel and S. Shankar and J. Steinhardt and I. Goodfellow and P. S. Liang and others},
  title     = {Enabling Certification of Verification-Agnostic Networks via Memory-Efficient Semidefinite Programming},
  booktitle = {Advances in Neural Information Processing Systems (NeurIPS)},
  year      = {2020}
}

@article{fazlyab2019efficient,
  title={Efficient and accurate estimation of lipschitz constants for deep neural networks},
  author={Fazlyab, Mahyar and Robey, Alexander and Hassani, Hamed and Morari, Manfred and Pappas, George},
  journal={Advances in neural information processing systems},
  volume={32},
  year={2019}
}

@article{wang2018efficient,
  title={Efficient formal safety analysis of neural networks},
  author={Wang, Shiqi and Pei, Kexin and Whitehouse, Justin and Yang, Junfeng and Jana, Suman},
  journal={Advances in neural information processing systems},
  volume={31},
  year={2018}
}

@inproceedings{chiu2025sdp,
  title={SDP-CROWN: Efficient Bound Propagation for Neural Network Verification with Tightness of Semidefinite Programming},
  author={Chiu, Hong-Ming and Chen, Hao and Zhang, Huan and Zhang, Richard Y},
  booktitle={International Conference on Machine Learning},
  pages={10449--10468},
  year={2025},
  organization={PMLR}
}

@inproceedings{zhou2025clip,
  title={Clip-and-Verify: Linear Constraint-Driven Domain Clipping for Accelerating Neural Network Verification},
  author={Zhou, Duo and Chavez, Jorge and Chen, Hesun and Hanasusanto, Grani A and Zhang, Huan},
  booktitle={The Thirty-ninth Annual Conference on Neural Information Processing Systems},
  year={2025}
}

@article{muller2022prima,
  title={PRIMA: general and precise neural network certification via scalable convex hull approximations},
  author={M{\"u}ller, Mark Niklas and Makarchuk, Gleb and Singh, Gagandeep and P{\"u}schel, Markus and Vechev, Martin},
  journal={Proceedings of the ACM on Programming Languages},
  volume={6},
  number={POPL},
  pages={1--33},
  year={2022},
  publisher={ACM New York, NY, USA}
}

@article{ma2024relu,
  title={Relu hull approximation},
  author={Ma, Zhongkui and Li, Jiaying and Bai, Guangdong},
  journal={Proceedings of the ACM on Programming Languages},
  volume={8},
  number={POPL},
  pages={2260--2287},
  year={2024},
  publisher={ACM New York, NY, USA}
}
